\documentclass[10pt,twocolumn,letterpaper]{article}

\usepackage[pagenumbers]{cvpr}             

\usepackage{subcaption}
\usepackage{multirow} 
\usepackage{bm}
\usepackage[table,dvipsnames]{xcolor}
\usepackage{colortbl}
\newcommand{\custombox}[3]{
  \setlength{\fboxsep}{#1} 
  \colorbox{#2}{#3} 
}
\usepackage{mathtools}
\usepackage{makecell}
\newcommand{\RNum}[1]{\uppercase\expandafter{\romannumeral #1\relax}}
\DeclareCaptionLabelFormat{withfigure}{Fig.\arabic{figure}#2}
\definecolor{mygray}{gray}{.9}
\definecolor{LightCyan}{rgb}{0.88,1,1}
\definecolor{lightcoral}{RGB}{240,128,128}
\definecolor{lightsalmon}{RGB}{255,160,122}
\definecolor{lightgray}{RGB}{128,128,128}
\definecolor{lightblue}{RGB}{212,239,251}
\definecolor{lightgreen}{RGB}{220,255,220}
\definecolor{lightpink}{RGB}{255,182,193}
\definecolor{cornsilk}{RGB}{255,248,220}
\definecolor{mintcream}{RGB}{245,255,250}
\definecolor{lavenderblush}{RGB}{255,240,245}
\definecolor{ghostwhite}{RGB}{248,248,255}
\definecolor{honeydew}{RGB}{240,255,240}
\definecolor{aliceblue}{RGB}{240,248,255}
\definecolor{ivory}{RGB}{255,255,240}
\definecolor{azure}{RGB}{240,255,255}
\definecolor{seashell}{RGB}{255,245,238}
\definecolor{textyellow}{RGB}{255,215,0}
\definecolor{gold}{RGB}{246, 228, 141}
\definecolor{silver}{RGB}{215, 215, 215}
\definecolor{bronze}{RGB}{229, 188, 151}
\definecolor{BrickRed}{RGB}{182, 50, 28}

\DeclareMathOperator*{\argmin}{\arg\min}

\definecolor{cvprblue}{rgb}{0.21,0.49,0.74}
\usepackage[pagebackref,breaklinks,colorlinks,citecolor=cvprblue]{hyperref}

\title{HeadGAP: Few-Shot 3D Head Avatar via Generalizable Gaussian Priors}

\author{
Xiaozheng Zheng\textsuperscript{\rm 1} 
\quad
Chao Wen\textsuperscript{\rm 1}$^{\dagger}$
\quad
Zhaohu Li\textsuperscript{\rm 1}
\quad
Weiyi Zhang\textsuperscript{\rm 1}
\quad
Zhuo Su\textsuperscript{\rm 1}
\quad
Xu Chang\textsuperscript{\rm 1}
\\
Yang Zhao\textsuperscript{\rm 1}
\quad
Zheng Lv\textsuperscript{\rm 1}
\quad
Xiaoyuan Zhang\textsuperscript{\rm 1}
\quad
Yongjie Zhang\textsuperscript{\rm 1}
\quad
Guidong Wang\textsuperscript{\rm 1}
\quad
Lan Xu\textsuperscript{\rm 2}
\vspace{5pt}\\
\textsuperscript{\rm 1}ByteDance
\qquad
\textsuperscript{\rm 2}ShanghaiTech University
}

\begin{document}
\twocolumn[{
\renewcommand\twocolumn[1][]{#1}
\maketitle

\begin{center}
    \vspace{-1.5em}
    \centering
    \captionsetup{type=figure}
    \vspace{-.7em}
    \includegraphics[width=0.95\textwidth ]{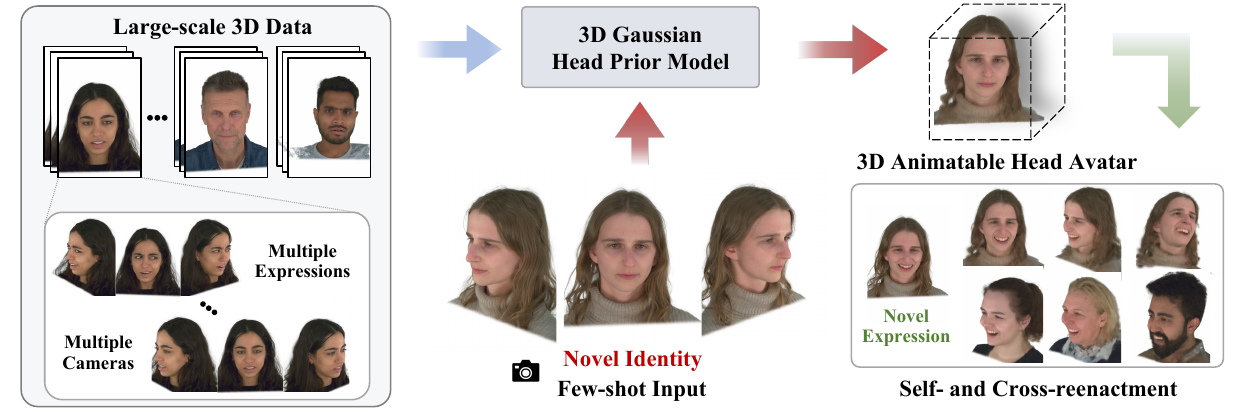}
        \captionof{figure}{We present HeadGAP to create photo-realistic animatable 3D head avatars from only a few or even one image of the target person. 
        \textcolor{NavyBlue}{\textbf{Firstly}}, we utilize large-scale 3D data to learn 3D head prior with our designed 3D Gaussian head prior model. 
        \textcolor{Mahogany}{\textbf{Secondly}}, we can use few-shot data to create 3D animatable avatars. 
        \textcolor{OliveGreen}{\textbf{Finally}}, we can animate the few-shot avatars with novel expressions.}
    \label{fig:teaser}
\end{center}
}]

{
  \renewcommand{\thefootnote}
    {\fnsymbol{footnote}}
  \footnotetext{Project page: \url{https://headgap.github.io/}} 
}

{
  \renewcommand{\thefootnote}
    {\fnsymbol{footnote}}
  \footnotetext{$^{\dagger}$ Corresponding author} 
}

\maketitle

\begin{abstract}
\vspace{-1em}
In this paper, we present a novel 3D head avatar creation approach capable of generalizing from few-shot in-the-wild data with high-fidelity and animatable robustness. Given the underconstrained nature of this problem, incorporating prior knowledge is essential. Therefore, we propose a framework comprising prior learning and avatar creation phases. The prior learning phase leverages 3D head priors derived from a large-scale multi-view dynamic dataset, and the avatar creation phase applies these priors for few-shot personalization. Our approach effectively captures these priors by utilizing a Gaussian Splatting-based auto-decoder network with part-based dynamic modeling. Our method employs identity-shared encoding with personalized latent codes for individual identities to learn the attributes of Gaussian primitives. During the avatar creation phase, we achieve fast head avatar personalization by leveraging inversion and fine-tuning strategies. Extensive experiments demonstrate that our model effectively exploits head priors and successfully generalizes them to few-shot personalization, achieving photo-realistic rendering quality, multi-view consistency, and stable animation.

\end{abstract}    
\section{Introduction}
\label{sec:intro}

Creating photo-realistic 3D avatars is a central challenge in computer graphics, encompassing applications such as movies, games, AR/VR, and the metaverse. There is a significant interest in generating digital avatars from real-world captures to create a precise digital copy of an actual person. These digital avatars can be animated and rendered from various viewpoints, maintaining high visual fidelity.

Recent advances~\cite{lombardi2021mixture,zielonka2023instant,xu2023avatarmav,kirschstein2024diffusionavatars,zheng2023pointavatar} have achieved photo-realistic rendering quality of digital humans.
In particular, 3D Gaussian Splatting (3DGS)~\cite{kerbl20233d} has been widely adopted for head avatars~\cite{qian2024gaussianavatars,xu2024gaussian,xiang2024flashavatar,giebenhain2024npga,chen2023monogaussianavatar,luo2024gaussianhair} due to its efficient and realistic rendering capabilities. 
However, these advancements rely heavily on publicly available multi-view~\cite{kirschstein2023nersemble} or sequential datasets~\cite{gafni2021dynamic,zheng2022avatar}, which are labor-intensive to capture and process for the average user.
To address this limitation, many studies~\cite{yu2024one2avatar,chu2024gpavatar,deng2024portrait4d,chen2024morphable,ding2023diffusionrig} aim to reduce the high data requirements for creating 3D avatars, allowing users to generate avatars from just a few images. 
Unfortunately, these methods often suffer from significant performance degradation compared to those~\cite{qian2024gaussianavatars,xu2024gaussian,giebenhain2024npga} that utilize dense data for avatar creation. 
The challenge of generating few-shot personalized head avatars with high fidelity and stable animation remains unresolved.

To this end, we propose \textbf{HeadGAP} to facilitate high-fidelity few-shot head avatar creation. 
As illustrated in \cref{fig:teaser}, the core of HeadGAP lies in learning \emph{generalizable} 3D Gaussian head priors from large-scale data and leveraging them to create \emph{high-quality} personalized head avatars with few-shot input.
As shown in \cref{fig:whole-pipeline}, the HeadGAP framework consists of two phases: 1) the \emph{prior learning} phase and 2) the \emph{few-shot personalization} phase.
In the \emph{prior learning} phase, multi-view dynamic data is used to embed 3D prior knowledge into \textbf{GAPNet} (\textbf{GA}ussian \textbf{P}rior \textbf{Net}work). 
The \emph{prior learning} phase is conducted only once. 
Subsequently, the \emph{few-shot personalization} phase uses the learned priors to create avatars of new identities by inversion and fine-tuning.
This framework focuses on two perspectives that would allow GAPNet to learn effective priors: 
1) To achieve \emph{high-quality}, we introduce a 3DGS-based head representation boosted with part-based and dynamic modeling.
2) For enhancing \emph{generalizability}, we design GAPNet in an auto-decoder manner, which constructs continuous part-based identity spaces that can serve as powerful generative priors to guide the few-shot creation.
Additionally, we leverage mesh tracking priors by predicting Gaussian attributes relative to the tracked mesh~\cite{qian2024gaussianavatars}.

We conduct comprehensive experiments on the NeRSemble dataset~\cite{kirschstein2023nersemble} to substantiate our design choices and demonstrate our method's superiority over existing approaches. To illustrate the robustness and practical applicability of our method, we present numerous avatars of novel identities generated from both public datasets and images captured with consumer-grade devices.

In summary, our contributions can be listed as follows:
\begin{itemize}
    \item We introduce a novel framework that exploits generalizable 3D Gaussian priors for fast 3D head avatar personalization using only a few input images. These avatars exhibit high fidelity and consistent animatable quality.
    \item We present proper designs that effectively utilize part-based dynamic Gaussian head priors and generalize them for high-quality few-shot head avatar personalization.
    \item We substantiate the efficacy and robustness of our framework through comprehensive experiments. Meanwhile, we showcase its potential in real scenarios by avatar creation using images captured by consumer-grade devices.
\end{itemize}

{
\setlength{\abovecaptionskip}{5pt plus 3pt minus 2pt} 
\begin{figure}[t!]
    \centering
    \includegraphics[width=0.5\textwidth]{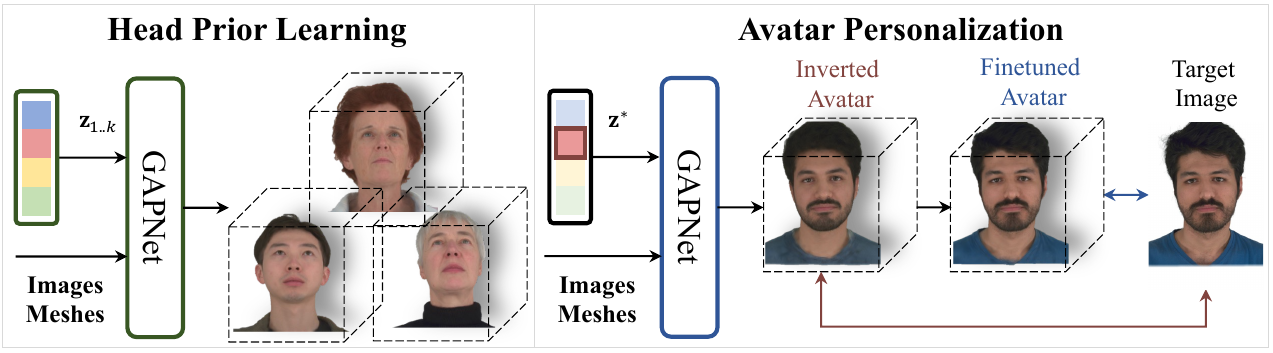}  
    \caption{
    HeadGAP framework. 
    The prior learning phase uses different IDs' data to embed head priors into the \textcolor{OliveGreen}{GAPNet}.
    The personalization phase firstly optimizes \textcolor{Mahogany}{identity codes} to obtain the inverted avatar, then updates the \textcolor{NavyBlue}{GAPNet} to get the fine-tuned avatar.
    }
    \label{fig:whole-pipeline}
    \vspace{-1.5em}
\end{figure}
}

{
\setlength{\abovecaptionskip}{5pt plus 3pt minus 2pt} 
\begin{figure*}[t]
    \centering
    \includegraphics[width=1.00\textwidth]{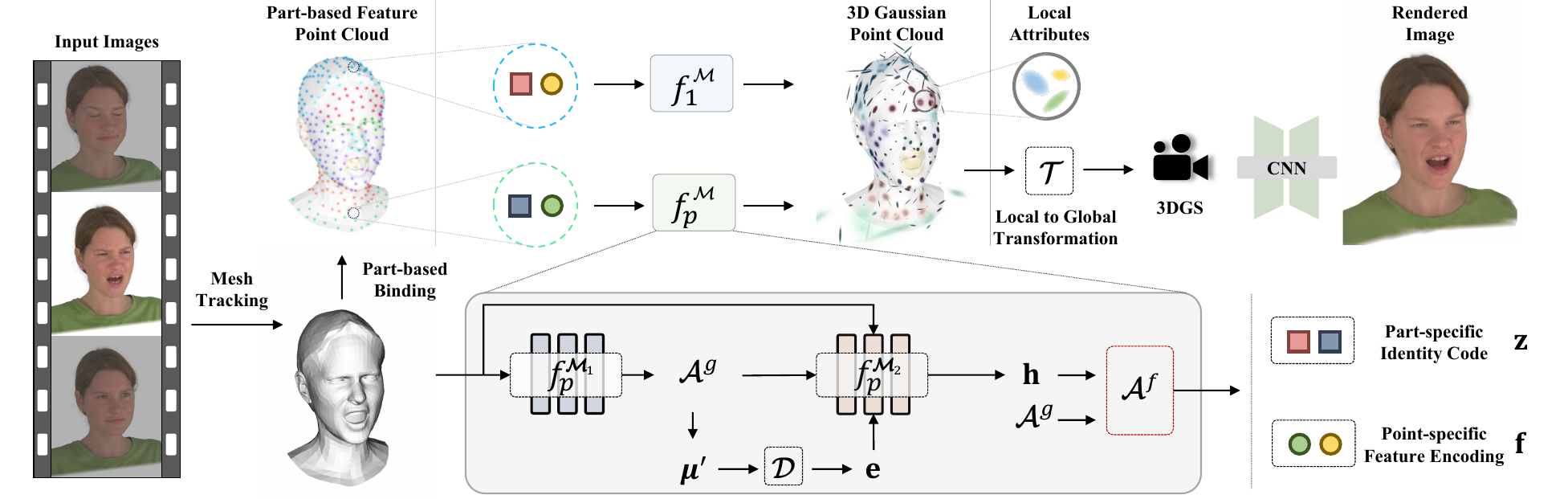}
    \caption{
    Illustration of the GAPNet. 
    Given the tracked meshes of the input images, GAPNet binds part-based Gaussian primitives with initialized features to the mesh. 
    Then, it employs part-specific modules to predict the local attributes of each primitive.
    The local attributes are transformed into global ones for 3DGS rendering.
    Finally, the renderings are fed into the CNN to obtain the final rendered images.
    }
    \label{fig:gapnet}
    \vspace{-1em}
\end{figure*}
}

\section{Related Work}
\label{sec:related}

\noindent
\textbf{3D Animatable Head Avatar.}
Since the advent of 3D neural implicit representations, remarkable progress has been made in creating animatable 3D head avatars from monocular or multi-view videos with various expressions and poses.
Existing works have explored varieties of avatar representation.
Some previous approaches~\cite{kim2018deep,ma2021pixel,grassal2022neural} employ 3DMM~\cite{li2017learning,blanz2023morphable} with neural textures.
Many recent studies~\cite{gafni2021dynamic, hong2022headnerf,gao2022reconstructing,lombardi2021mixture,zielonka2023instant,bai2023learning,athar2022rignerf,zheng2022avatar,xu2023avatarmav,kirschstein2024diffusionavatars} focus on creating neural volumetric avatars.
3DMM~\cite{blanz2003face,paysan20093d,li2017learning,blanz2023morphable,giebenhain2023learning} is often employed in those approaches.
More recently, point-based representations are widely adopted~\cite{zheng2023pointavatar,qian2024gaussianavatars,shao2024splattingavatar,xu2024gaussian,xiang2024flashavatar,chen2023monogaussianavatar,giebenhain2024npga,giebenhain2024mononphm}.
Among those works, 3DGS~\cite{kerbl20233d} is the most prevalent representation due to its efficient rendering and topological flexibility.
Similar to these approaches, our approach is also based on 3DGS. 
However, there exist significant differences between previous works and ours, including our model being designed with 1) \emph{part-based dynamic modeling} and designed for 2) \emph{3DGS-based generative modeling} rather than single-subject modeling.

\noindent
\textbf{One-shot 2D Head Avatar.}
One-shot 2D head avatar synthesis has attracted lots of attention in recent years.
Plenty of works leverage 2D generative models for talking head synthesis at high fidelity. 
One part of those works~\cite{wiles2018x2face,siarohin2019first,ren2021pirenderer,wang2021one,hong2022depth,drobyshev2022megaportraits,gao2023high,zhang2023metaportrait} learn latent deformed features and feed them to 2D generators for face reenactment. 
Some other studies
\cite{yin2022styleheat,bounareli2023hyperreenact,guan2023stylesync} map images to the latent space of a pre-trained StyleGAN2~\cite{karras2020analyzing}. 
While 2D-based methods can produce photorealistic images, they struggle to preserve the 3D consistency. 
Therefore, several methods \cite{hong2022headnerf,khakhulin2022realistic,li2023one,li2024generalizable,ma2023otavatar,yu2023nofa,chu2024gpavatar,ye2024real3d,ma2024cvthead,deng2024portrait4d,deng2024portrait4dv2} pursue animatable 3D head synthesis. 
They often resort to monocular 3DMM~\cite{deng2019accurate,feng2021learning} for providing geometry or pose guidance. 
Another line of work for 3D-aware portrait generations is also capable of few-shot avatar animations.
Lots of studies~\cite{nguyen2019hologan,schwarz2020graf,gu2021stylenerf,chan2021pi,chan2022efficient,deng2022gram,chen2023mimic3d,sun2022ide} demonstrate that the combination of 3D representations and adversarial learning on monocular images makes it possible to learn a 3D-aware generator for multi-view image generation.
Many works~\cite{bergman2022generative,sun2023next3d,sun2022controllable,tang20233dfaceshop,wu2022anifacegan,wu2023aniportraitgan,xu2023omniavatar} introduce 3DMM for animation control. 
These methods can be combined with advanced GAN inversion techniques~\cite{deng2023learning,ko20233d,roich2022pivotal,xie2023high,yin2023nerfinvertor} for head avatar reconstruction.
However, those 2D-based approaches are still inferior in 3D consistency due to their representations and training schemes.

\noindent
\textbf{Few-shot Head Avatar with Data-driven 3D Priors}.
We focus on the few-shot 3D avatar personalization with data-driven 3D priors from large-scale data. 
To solve this problem, there are also some recent works~\cite{cao2022authentic, buhler2023preface,yang2024vrmm,yu2024one2avatar,chen2024morphable} designed in this manner.
Morphable Diffusion~\cite{chen2024morphable} introduces a multi-view consistent diffusion model to create head avatars from a single image.
Preface~\cite{buhler2023preface} trains a NeRF-based auto-decoder generative model and achieves few-shot high-fidelity 3D static head creations. 
PhoneScan~\cite{cao2022authentic} extends MVP~\cite{lombardi2021mixture} to an auto-encoder generative model and supports novel avatar creation with the phone-captured data.
VRMM~\cite{yang2024vrmm} is an auto-decoder generative model built upon MVP~\cite{lombardi2021mixture}, which supports few-shot relightable avatar creation.
One2Avatar~\cite{yu2024one2avatar} adapts MonoAvatar~\cite{bai2023learning} to an auto-decoder generative model based on a 3DMM-anchored neural radiance field~\cite{mildenhall2021nerf}.
Similar to \cite{buhler2023preface,yu2024one2avatar,yang2024vrmm}, we also design our model in an auto-decoder manner.
Different from these approaches relying on volume rendering~\cite{cao2022authentic, buhler2023preface,yang2024vrmm,yu2024one2avatar} or diffusion model~\cite{chen2024morphable}, our approach employs 3DGS for rendering.
We concentrate on designing a 3DGS-based generative model to achieve high-fidelity few-shot personalization with robust animations.

\setlength{\abovedisplayskip}{4pt} 
\setlength{\belowdisplayskip}{4pt}
\setlength{\abovedisplayshortskip}{4pt}
\setlength{\belowdisplayshortskip}{4pt}

\section{Method}
\label{sec:method}
In this section, we first introduce the preliminary (\cref{sec:preliminary}).
Then, we detail our avatar representation designed for creating head avatars with learned generalizable head prior knowledge (\cref{sec:avatar representation}).
Finally, we present our HeadGAP framework (\cref{fig:whole-pipeline}), including 1) head prior learning phase (\cref{sec:prior learning}) and 2) few-shot personalization phase (\cref{sec:few-shot personalization}).

\subsection{Preliminary}
\label{sec:preliminary}
\textbf{3D Gaussian Splatting (3DGS)~\cite{kerbl20233d}}
proposes a point-based scene representation, where each point represents a Gaussian primitive that is described by a global space position $\bm{\mu}$, rotation $\mathbf{r}$, scale $\mathbf{S}$, opacity $\alpha$ and color $\mathbf{c}$.
In the following, we let the following notation:
\begin{equation}
\label{eq:3dgs}
\begin{array}{c}
    \mathcal{A} = \{ \bm{\mu}, \mathbf{r}, \mathbf{S}, \alpha, \mathbf{c}\}, 
    \quad 
    \mathbf{I} = \mathcal{R}(\mathcal{A}, \pi_{\mathbf{K},\mathbf{E}})
\end{array}
\end{equation}
denote the set of attributes $\mathcal{A}$ composing the Gaussian point cloud, and its tile-based differentiable rasterization $\mathcal{R}$ into an image $\mathbf{I}$ under the camera projection $\pi$ described by intrinsic and extrinsic parameters $\mathbf{K}$ and $\mathbf{E}$ respectively.

\noindent
\textbf{GaussianAvatars}~\cite{qian2024gaussianavatars} connects Gaussian primitives to the mesh faces.
For each primitive, the position $\bm{\mu}^{\prime}$, rotation $\mathbf{r}^{\prime}$, and scaling $\mathbf{S}^{\prime}$ are initialized in the local space. 
During rendering, these properties are converted into the global space:
\begin{equation}
\label{eq:gaussianavatars}
\begin{array}{c}
    \mathbf{r} = \mathbf{R}\mathbf{r}^{\prime}, \quad
    \bm{\mu} = s\mathbf{R}\bm{\mu}^{\prime} + \mathbf{T},\quad
    \mathbf{S} = s\mathbf{S}^{\prime}, 
\end{array}
\end{equation}
where $\mathbf{R}$ describes the orientation of the triangle face in the global space, 
$s$ describes the scaling, 
and $\mathbf{T}$ describes the mean position of three vertices of a triangle face.
For simplicity, we define $\mathcal{T}$ as the operation that transforms the local Gaussian attributes $\mathcal{A}^{\prime}$ to global ones:
\begin{equation}
\label{eq:gaussianavatars2}
\mathcal{A} = \mathcal{T}(\mathcal{A}^{\prime},\mathcal{M}),  \quad \mathcal{M} = \{ \bm{\beta}, \bm{\theta}, \bm{\phi}, \bm{\delta} \},
\end{equation}
where $\mathcal{M}$ denotes the input FLAME parameters, consisting of shape parameters $\bm{\beta}$, pose parameters $\bm{\theta}$, expression parameters $\bm{\phi}$, and static vertex offsets $\bm{\delta}$.

\subsection{Avatar Representation}
\label{sec:avatar representation}
Our avatar representation is based on an auto-decoder prior model~\cite{buhler2023preface,zheng2024ohta,yang2024vrmm,yu2024one2avatar} that can learn head prior knowledge from multiple identities and be used for head avatar creation from few-shot images.
As illustrated in Fig.~\ref{fig:gapnet}, our representation builds upon a point-based representation with \emph{part-based modeling}, where each point is only responsible for one semantic part.
Firstly, we initialize the \emph{part-based feature point cloud} consisting of 1) \emph{part-based identity code} and 2) \emph{point-specific feature encoding}, based on the tracked mesh. 
Then, we conduct \emph{dynamic modeling} by feeding the feature point cloud to the \emph{part-based multi-layer perceptions} (MLPs) to regress the Gaussian attributes of all the points for 3DGS rendering. 
Finally, we utilize a \emph{convolutional neural network} (CNN) module to refine the 3DGS renderings to obtain the final rendered image.
In the following, we will describe those key components.

\noindent
\textbf{Part-based Feature Point Cloud.}
For initializing the \emph{part-based feature point cloud} based on the tracked mesh $\mathcal{M}$, we first utilize UV-based initialization~\cite{xiang2024flashavatar} to obtain the point cloud with $n$ Gaussian primitives, with each pixel in the UV map bound to one triangle of the mesh.
The initialization contributes to more uniform primitives distributed on the head region than face-based initialization~\cite{qian2024gaussianavatars}.

Then, we set up the initial features for the point cloud.
The features contains two types, including the 1) \emph{point-specific feature encodings} $\mathbf{f} = \{\mathbf{f}_{i} \in \mathbb{R}^{c_{1}}\}_{i=1}^{n}$ and 2) \emph{part-specific identity codes} $ \mathbf{z} = \{ \{ \mathbf{z}_{j}^{l} \in \mathbb{R}^{c_{2}} \}^{p}_{l=1} \} ^{k}_{j=1}$, where $p$ and $k$ denote the part and identity number respectively.
The point encodings $\mathbf{f}$ embeds identity-shared priors and the identity codes $\mathbf{z}$ serve as the identity codebook for the auto-decoder model. 
All the encodings are randomly initialized learnable parameters.
The part of a primitive is determined by its parent triangle and 
the identity codes are the same for all the primitives belonging to the same part.

\noindent
\textbf{Part-based Dynamic Gaussian Attributes Modeling.}
For simplicity, we use $\mathbf{f}$ and $\mathbf{z}$ to denote the per-point features belonging to a specific part $p$ and also omit the part notation for other notations, unless otherwise stated.
Given $\mathbf{f}$ and $\mathbf{z}$, we regress dynamic local Gaussian attributes by: 
\begin{equation}
\label{eq:head-representation}
\begin{array}{c}
    \mathcal{A}^{g} = f^{\mathcal{M}_{1}}_{p}(\mathbf{f}, \mathbf{z}), 
    \quad
    \mathbf{h} = f^{\mathcal{M}_{2}}_{p}(\mathbf{f}, \mathbf{z}, \mathbf{e}, \mathcal{A}^{g}),
\end{array}
\end{equation}
where $\mathbf{h}$ is point appearance attribute, 
$\mathcal{A}^{g} = \{ \bm{\mu}^{\prime}, \mathbf{r}^{\prime}, \mathbf{S}^{\prime}, \alpha\}$ denotes other attributes,
and $\mathbf{e} \coloneqq \mathcal{D}(\bm{\mu}^{\prime}) = \mathcal{T}(\bm{\mu}^{\prime}) - \mathcal{T}(\bm{\mu}^{\prime}_{neutral})$ is the point-specific dynamic signals obtained by subtracting the global neutral point position $\mathcal{T}(\bm{\mu}^{\prime}_{neutral})$ from the global posed point position $\mathcal{T}(\bm{\mu}^{\prime})$.
Both $f^{\mathcal{M}_{1}}_{p}$ and $f^{\mathcal{M}_{2}}_{p}$ are part-specific MLPs.
We define the overall dynamic modeling as:
$\mathcal{A}^{f} = f^{\mathcal{M}}_{p}(\mathbf{f},\mathbf{z})$
, where $\mathcal{A}^{f} = \mathcal{A}^{g}  \cup \{\mathbf{h} \}$ denotes the final Gaussian attributes used for splatting.

Part-based and dynamic modeling contribute to better few-shot performance, as shown in \cref{sec:model-analysis}.
The part-based modeling allows the specialized module to learn the particular part's priors, resulting in easier optimization and more powerful priors.
The dynamic modeling employs point-specific expression signals $\mathbf{e}$ for predicting dynamic local attributes, which is better at capturing dynamic details than GaussianAvatars~\cite{qian2024gaussianavatars} using static local attributes.

\noindent
\textbf{Gaussian Splatting with CNN refinement.}
Inspired by recent works~\cite{xu2023latentavatar,xu2024gaussian}, we apply a screen-space CNN $f^{\mathcal{C}}$ to refine the rendered results:
\begin{align}
\label{eq:head-representation-cnn}
    [\mathbf{I}_{rgb}, \mathbf{I}_{h}] &= \mathcal{R}(\mathcal{T}(\mathcal{A}_{f}, \mathcal{M}), \pi_{\mathbf{K},\mathbf{E}}),\\
    \mathbf{I} &= f^{\mathcal{C}}([\mathbf{I}_{rgb}, \mathbf{I}_{h}]),
\end{align}
where $\mathcal{A}^{f} = \{ \mathcal{A}^{f}_{i}\}_{i=1}^{n}$ denotes the final Gaussian attributes for all the points, $\mathbf{I}_{rgb}$ denotes rendered RGB images, and $\mathbf{I}_{h}$ is a latent feature image used for the CNN refinement.

Different from previous methods~\cite{xu2023latentavatar,xu2024gaussian}, we do not conduct super-resolution, but keep the input and output with the same resolution for refinement.
We aim to use large-scale data to enable CNN to capture generalizable structured appearance priors that are challenging to exploit by our 3DGS-based representation.
As indicated in \cref{sec:model-analysis}, using CNN-based refinement can indeed capture those priors to render more photo-realistic results for few-shot personalization.

\noindent
\textbf{Overall Representation.}
The overall head avatar representation $\mathcal{H}$ is defined formally:
\begin{equation}
\label{eq:mapping}
\begin{array}{c}
    \mathcal{H}: (\mathcal{M}; f^{\mathcal{M}}, f^{\mathcal{C}}, \mathbf{f}, \mathbf{z}) \mapsto \mathbf{I}\;, \\
\end{array}
\end{equation}
where $f^{\mathcal{M}}=\{f^{\mathcal{M}}_{l}\}_{l=1}^{p}$ denotes the MLPs for all parts.

\begin{table*}[h!]
  \centering
  \resizebox{0.95\linewidth}{!}{
  \begin{tabular}{cccccccccc}
    \toprule 
    \multirow{2}{*}{Method} & \multirow{2}{*}{Reference} & \multirow{2}{*}{Input} & \multicolumn{3}{c}{\cellcolor{honeydew!200}Frontal view} & \multicolumn{4}{c}{\cellcolor{ghostwhite!200} All views} \\
    & & & \cellcolor{honeydew!200}LPIPS$\downarrow$ & \cellcolor{honeydew!200}PSNR$\uparrow$ & \cellcolor{honeydew!200}SSIM$\uparrow$ & \cellcolor{ghostwhite!200}LPIPS$\downarrow$ & \cellcolor{ghostwhite!200}PSNR$\uparrow$ & \cellcolor{ghostwhite!200}SSIM$\uparrow$ &
    \cellcolor{ghostwhite!200}ID$\uparrow$\\
    \midrule
    ROME~\cite{khakhulin2022realistic}  & ECCV'22 & 1 image & 0.237 & 17.56 & \custombox{2pt}{bronze}{0.813}  & \custombox{2pt}{bronze}{0.286} & \custombox{2pt}{gold}{15.45} & \custombox{2pt}{gold}{0.796} & 0.658$\pm$0.119 \\
    GOHA~\cite{li2024generalizable} &  NeurIPS'23 & 1 image & 0.224  & 15.85  & 0.760 & 0.308 & 13.13 & 0.736 & 0.588$\pm$0.146 \\
    VOODOO3D~\cite{tran2024voodoo} & CVPR'24 & 1 image & 0.294 & 17.07  & 0.773  & 0.310 & \custombox{2pt}{bronze}{14.59} & 0.752 & 0.626$\pm$0.111  \\
    HiDe-NeRF~\cite{li2023one} & CVPR'23 & 1 image & 0.290  & 15.63  & 0.789 & 0.368 & 14.51 & \custombox{2pt}{bronze}{0.784} & 0.640$\pm$0.156  \\
    Portrait4Dv1~\cite{deng2024portrait4d} & CVPR'24 & 1 image & \custombox{2pt}{bronze}{0.180}  & 16.59  & 0.796  & 0.273  & 14.52 & 0.752 & \custombox{2pt}{bronze}{0.674$\pm$0.143}  \\
    Portrait4Dv2~\cite{deng2024portrait4dv2} & ECCV'24 & 1 image & \custombox{2pt}{silver}{0.155}  & \custombox{2pt}{bronze}{17.77}  & 0.810  & \custombox{2pt}{silver}{0.269} & 14.53 & 0.757 & \custombox{2pt}{silver}{0.694$\pm$0.141} \\
    GPAvatar~\cite{chu2024gpavatar} & ICLR'24 & 1 image & \custombox{2pt}{bronze}{0.180} & \custombox{2pt}{gold}{18.42} & \custombox{2pt}{silver}{0.827} & 0.294 & 13.83 & 0.775 & 0.631$\pm$0.169 \\
    
    Ours-SV & & 1 image & \custombox{2pt}{gold}{0.142} & \custombox{2pt}{silver}{17.91} & \custombox{2pt}{gold}{0.829} & \custombox{2pt}{gold}{0.217} & \custombox{2pt}{silver}{14.75} & \custombox{2pt}{silver}{0.792} & \custombox{2pt}{gold}{0.768$\pm$0.113} \\
    
    \midrule
    DiffusionRig~\cite{ding2023diffusionrig} & CVPR'23 & 20 images & 0.220 & 16.94 & 0.811  & 0.298 & 14.88 & 0.786 & \custombox{2pt}{silver}{0.817$\pm$0.112} \\
    NHA$^\dagger$~\cite{grassal2022neural} & CVPR'22 & mono video & 0.161 & 17.42 & 0.850 & \custombox{2pt}{silver}{0.266} & 14.77 & \custombox{2pt}{silver}{0.807} & \custombox{2pt}{bronze}{0.577$\pm$0.138} \\
    FlashAvatar$^\dagger$~\cite{xiang2024flashavatar} & CVPR'24 & mono video & \custombox{2pt}{silver}{0.146}  & \custombox{2pt}{bronze}{18.89}  & \custombox{2pt}{silver}{0.854}  & \custombox{2pt}{bronze}{0.286} & \custombox{2pt}{bronze}{16.56} & \custombox{2pt}{bronze}{0.791} & -- \\
    GaussianAvatars$^\ddagger$~\cite{qian2024gaussianavatars} & CVPR'24 &  3 images  & 0.320  & 16.19  & 0.723  & 0.337 & 15.80 & 0.705 & -- \\
    GaussianAvatars$^\blacklozenge$~\cite{qian2024gaussianavatars} & CVPR'24 &  3 videos & \custombox{2pt}{bronze}{0.147} & \custombox{2pt}{silver}{21.05} & \custombox{2pt}{bronze}{0.852} & 0.301 & \custombox{2pt}{silver}{16.62} & 0.728 & -- \\
    Ours  & & 3 images & \custombox{2pt}{gold}{0.138} & \custombox{2pt}{gold}{21.67} & \custombox{2pt}{gold}{0.866} & \custombox{2pt}{gold}{0.144} & \custombox{2pt}{gold}{20.90} & \custombox{2pt}{gold}{0.868} & \custombox{2pt}{gold}{0.821$\pm$0.094} \\

    \bottomrule
    
  \end{tabular}
  }
  \caption{Quantitative comparisons with state-of-the-art methods on NeRSemble~\cite{kirschstein2023nersemble} dataset. We use colors to denote the \custombox{1pt}{gold}{first}, \custombox{1pt}{silver}{second} and \custombox{1pt}{bronze}{third} places respectively. The results are averaged across novel-view and novel-pose.
  }
  \label{tab:Quantitative comparisons with SOTAs on NeRSemble}
  \vspace{-1em}
\end{table*}

\subsection{Head Prior Learning}
\label{sec:prior learning}
We highly rely on the head's prior knowledge to achieve high-fidelity avatar creation for the unconstrained problem with only a few input images. 
Among various priors, we aim at learning high-quality, animatable, and 3D-consistent head priors from available multi-view dynamic head data with multiple identities.
Therefore, the goal for the prior learning stage is to learn $k$ head avatars within the GAPNet, with the respective identity codes $\mathbf{z}_{1...k}$ and other network parameters optimized.
Before starting model training, we first conduct FLAME tracking for the training data to obtain $\mathcal{M}$.
Then, we use those data to jointly optimize $\mathcal{M}$, $f^{\mathcal{M}}$, ${f^{\mathcal{C}}}$, $\mathbf{f}$, and $\mathbf{z}$ with our total loss term: 
\begin{align}
\label{eq:loss}
    \mathcal{L} = &\mathcal{L}_{rec}(\mathbf{I}, \mathbf{I}^{*}) + \mathcal{L}_{rec}(\mathbf{I}_{rgb}, \mathbf{I}^{*}) + \notag \\
    &\lambda_{m} \mathcal{L}_{rec}(\mathbf{I}_{m}, \mathbf{I}^{*}_{m}) + \mathcal{L}_{reg},
\end{align}
where $\mathcal{L}_{rec}$ and $\mathcal{L}_{reg}$ denote the image reconstruction loss and training regularization loss respectively.
The ground truth image is denoted as $\mathbf{I}^{*}$.
To improve the fidelity of the mouth region, we further supervise mouth region $\mathbf{I}_{m}$ with masked ground truth mouth region $\mathbf{I}^{*}_{m}$, inspired by FlashAvatar~\cite{xiang2024flashavatar}.
Specifically, the image reconstruction loss:
\begin{equation}
\label{eq:reconstruction loss}
\begin{array}{c}
    \mathcal{L}_{rec} = \lambda_{l1}\mathcal{L}_{l1} + \lambda_{ssim}\mathcal{L}_{ssim} + \lambda_{lpips}\mathcal{L}_{lpips}
\end{array}
\end{equation}
consists of L1 loss $\mathcal{L}_{l1}$, SSIM loss $\mathcal{L}_{ssim}$, and perceptual loss $\mathcal{L}_{lpips}$ with the VGG as the backbone. 
Meanwhile, the training regularization loss:
\begin{equation}
\label{eq:regularization loss}
\begin{array}{c}
    \mathcal{L}_{reg} = \lambda_{\alpha}\mathcal{L}_{\alpha} + \lambda_{s}\mathcal{L}_{s} + \lambda_{\mu}\mathcal{L}_{\mu} + \lambda_{arap}\mathcal{L}_{arap},
\end{array}
\end{equation}
includes opacity regularization $\mathcal{L}_{\alpha} = \| \mathbf{I}_{\alpha} - \tilde{\mathbf{I}}_{mask} \|_{1}$, primitive local scaling regularization $\mathcal{L}_{s} = \| \text{max}(\mathbf{s}, \epsilon_{s}) \|_{2}$,  primitive local position regularization $\mathcal{L}_{\mu} = \| \text{max}(\bm{\mu}, \epsilon_{\mu}) \|_{2}$, and ARAP (As-Rigid-As-Possible) regularization $\mathcal{L}_{arap}$ \cite{sorkine2007rigid}.
The opacity regularization $\mathcal{L}_{\alpha}$ is used to constrain the Gaussian primitives to stay within the head region and their opacity accumulated to $1$, which is computed between accumulated opacity image $\mathbf{I}_{\alpha}$ and the head mask $\tilde{\mathbf{I}}_{mask}$.
We utilize the same thresholds $\epsilon_{s}=0.6$ and $\epsilon_{\mu}=1$ for $\mathcal{L}_{s}$ and $\mathcal{L}_{\mu}$ respectively as \cite{qian2024gaussianavatars} to constraint the local scaling and position of the Gaussian primitives.
The ARAP regularization $\mathcal{L}_{arap}$ is employed for regularizing the optimization of static offset.
All the $\lambda$s described above are used for balancing different loss terms.

\subsection{Few-shot Personalization} 
\label{sec:few-shot personalization}
After the prior learning phase, we encode dynamic head prior knowledge within GAPNet. Consequently, all the learned parameters of GAPNet can serve as powerful priors to aid few-shot or even one-shot personalization.

Prior to personalization, we employ a tracker to acquire the FLAME~\cite{li2017learning} parameters of the input image.
Given input images with FLAME trackings, we first find the most similar avatar from the identity codebook through inversion.
Specifically, we optimize part-specific linear combination weights $\mathbf{w} \in \mathbb{R}^{k \times p \times 1}$ to obtain the identity code $\mathbf{z}^{*} = \text{softmax}(\mathbf{w}) \odot \mathbf{z} \in \mathbb{R}^{k \times p \times c_{2}}$ used for rendering an avatar similar to the input.
During the inversion optimization, we keep all the parameters of the network frozen except for $\mathbf{w}$.
Formally, given an input image $\mathbf{I}^{*}$ of the target identity, we optimize to render an image $\mathbf{I}$ that resembles the target identity.
This procedure is optimized with the loss function in \cref{eq:loss} with respect to $\mathbf{w}$.

Then, we start fine-tuning to update the network's parameters so that the avatar can capture the details of the target identity from the inputs.
We leverage prior knowledge in this procedure through three strategies.
First, we use small learning rates for all parameters except $\mathbf{f}$. 
Next, we exploit extracted part-based priors by excluding the fine-tuning for the mouth region, as modeling the highly flexible mouth region with few inputs is challenging.
Finally, we apply view regularization to prevent overfitting to the target view, inspired by previous methods~\cite{zheng2024ohta,roich2022pivotal}. 
Specifically, we constraint the fine-tuning results of some reference views with neutral face $\{\mathbf{R}_{i}\}_{i=1}^{m}$ to be close to the rendering results before fine-tuning $\{\tilde{\mathbf{R}}_{i}\}_{i=1}^{m}$, where $m$ is the number of the generated reference views.
With the prior knowledge, our personalized avatar achieves stable reenactment while preserving the details of the target identity.
The fine-tuning is conducted by minimizing the loss function in \cref{eq:loss}:
\begin{equation}
\label{eq:inverse}
\argmin_{\xi} \mathcal{L}_{f} = \mathcal{L}(\mathbf{I}, \mathbf{I}^{*}) + \lambda_{ref} \sum_{i=1}^{m}(\mathcal{L}(\mathbf{R}_{i}, \tilde{\mathbf{R}}_{i})),
\end{equation}
where $\xi$ denotes all the learnable parameters, and $\lambda_{ref}$ is used to balance different loss terms. 
\section{Experiments}
\label{sec:experiment}

\begin{figure*}[h]
    \centering
    \includegraphics[width=\textwidth]{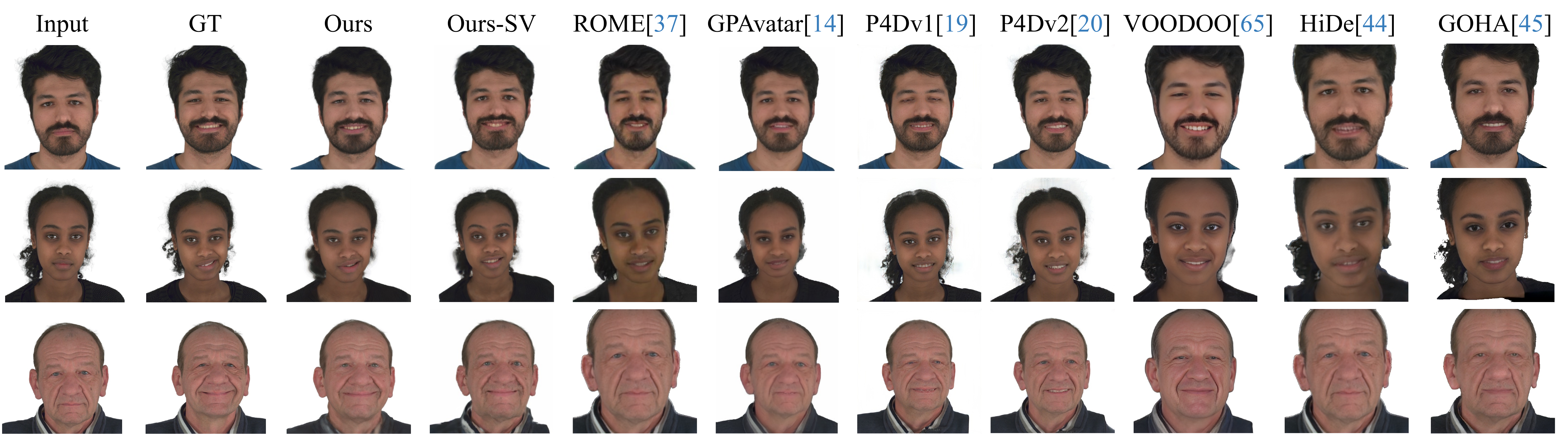}
    \caption{
    Qualitative comparisons of our approach against state-of-the-art methods using a single image as input.
    }
    \label{fig:exp-sota-single-view}
    \vspace{-1.5em}
\end{figure*}

\subsection{Setup}
\label{sec:setup}
\noindent{\textbf{Dataset.}}~
We utilize facial images of 164 subjects with 16 camera viewpoints in the NeRSemble~\cite{kirschstein2023nersemble} dataset for experiments.
We separated the data into training and testing sets, comprising 119 and 45 subjects respectively.
The training sets are used for prior learning, while the testing sets are employed to quantitatively and qualitatively evaluate few-shot personalization performance.
We also constructed an in-house dataset as part of the testing data.
In addition to the leave-out testing data, we conduct experiments on data captured by consumer-grade devices to evaluate the performance towards in-the-wild inputs.

\noindent{\textbf{FLAME Tracking.}}
Inspired by previous studies~\cite{qian2024gaussianavatars,zielonka2022towards}, we designed a tracking algorithm that can optimize the FLAME~ \cite{li2017learning} parameters with different numbers of input views. In the following experimental section, unless otherwise stated, the training data uses this tracking algorithm to obtain the ground truth 3DMM parameters. For the test data, we simplify the tracking to use sparse view inputs and provide the corresponding 3DMM parameters for testing.

\subsection{Implementation Details}
\label{sec:implementation}

\noindent{\textbf{Model Detail.}}
We divide the primitives into $p=11$ parts according to the face masks from FLAME \cite{li2017learning}.
We utilize $k=119$ identities for prior learning.
All the MLPs $f^{\mathcal{M}}$ consist of $4$ layers and the CNN $f^{\mathcal{C}}$ contains $6$ layers.

\noindent{\textbf{Training Detail.}}
We adopt Adam~\cite{kingma2014adam} optimizer for the training.
For prior learning, we set the batch size to 32. All parameters start with a learning rate of $1e^{-3}$, which decreases using a cosine scheduler. 
The prior model is trained on 8 A100 GPUs for 100K steps, taking around 2 days.
For few-shot personalization, we use a batch size of 1. 
Both inversion and fine-tuning take 500 steps, totaling about 5 minutes on an A100 GPU.
Please refer to supplementary materials for more details.

\vspace{-0.5em}
\subsection{Baselines and Metrics}
\label{sec:baselines}

\noindent
\textbf{Baselines.}
We classify baselines into two types based on their training approaches. The \textcolor{BrickRed}{Type-\RNum{1}}, like ours, uses multi-ID datasets to train prior features of the head, which can generalize to novel IDs, termed 
\textit{prior-based methods}.
The \textcolor{OliveGreen}{Type-\RNum{2}} requires individual training for each person, termed \textit{per-subject optimization methods}.
For \textcolor{BrickRed}{Type-\RNum{1}}, due to the lack of available code for most head avatar generation methods using a few views~\cite{yu2024one2avatar,yang2024vrmm,cao2022authentic}, we compare our method with approaches using a single image, including mesh-based methods~\cite{khakhulin2022realistic} and the state-of-the-art tri-plane based methods~\cite{li2024generalizable,tran2024voodoo,li2023one,deng2024portrait4d,deng2024portrait4dv2,chu2024gpavatar}. We also compare with 3D-aware diffusion models~\cite{ding2023diffusionrig} equipped with multi-view inputs.
For \textcolor{OliveGreen}{Type-\RNum{2}}, we compare our approach with methods using Gaussian Splatting~\cite{xiang2024flashavatar,qian2024gaussianavatars} or explicit mesh~\cite{grassal2022neural} as 3D representations. Per-subject methods require more training data, so we provide these baselines with monocular video$^\dagger$, multi-view images$^\ddagger$, or multi-view videos$^\blacklozenge$.

\noindent
\textbf{Metrics.}
We employ standard image quality metrics for our quantitative evaluations: 1) Peak Signal-to-Noise Ratio (PSNR), 2) Structure Similarity Index (SSIM), and 3) Learned Perceptual Image Patch Similarity (LPIPS), following previous works~\cite{qian2024gaussianavatars,yu2024one2avatar,gafni2021dynamic}.
Furthermore, we also report 4) ID that measures the identity similarity \cite{deng2019arcface} between the predictions and ground truth ones.

\begin{figure}[h!]
    \centering
    \includegraphics[width=0.98\columnwidth]{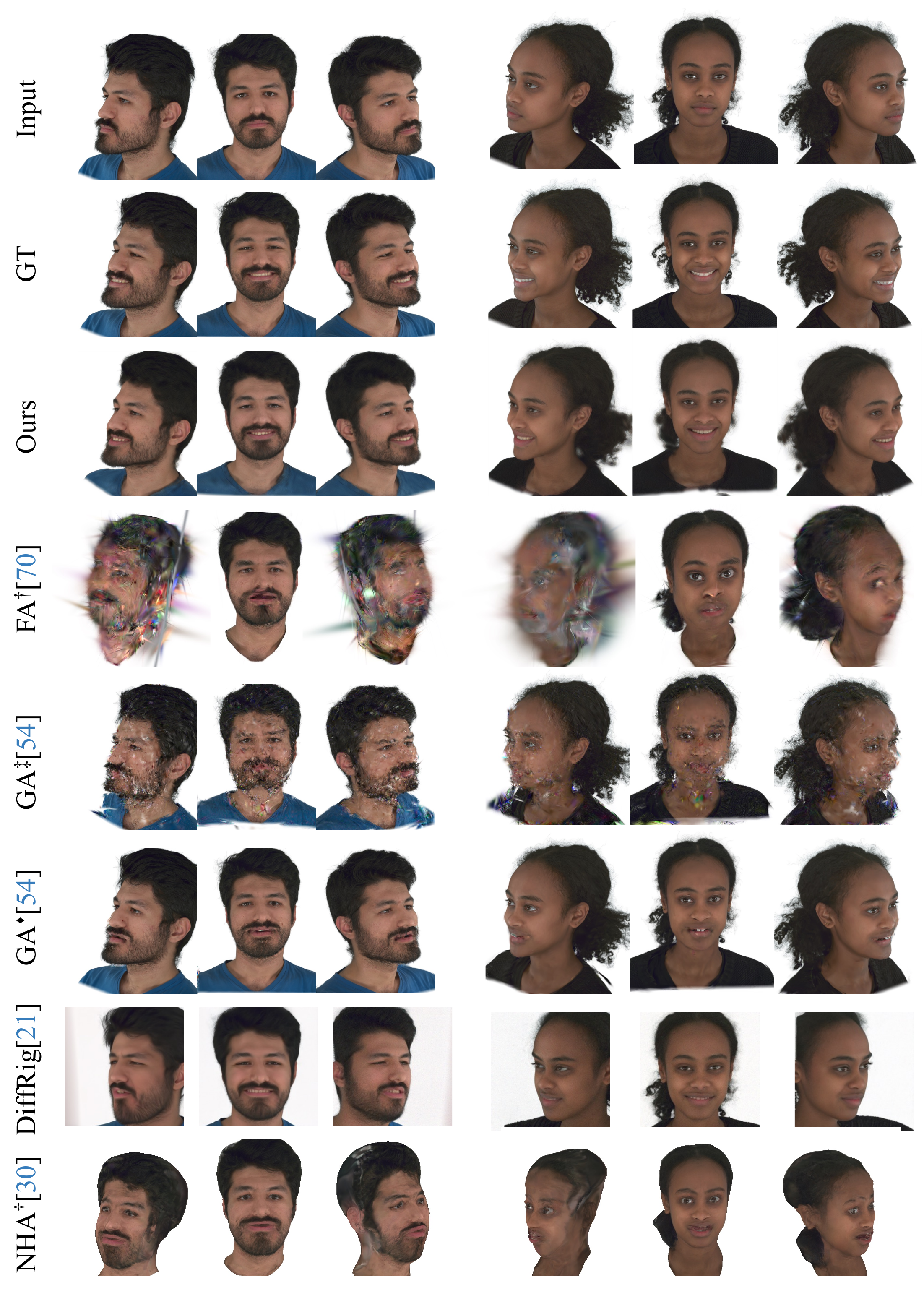}
    \caption{
    Qualitative comparisons of our approach against state-of-the-art methods using few-shot input.
    }
    \label{fig:exp-sota-all-view}
    \vspace{-1.5em}
\end{figure}

\begin{figure*}[h!]
    \centering
    \includegraphics[width=\textwidth]{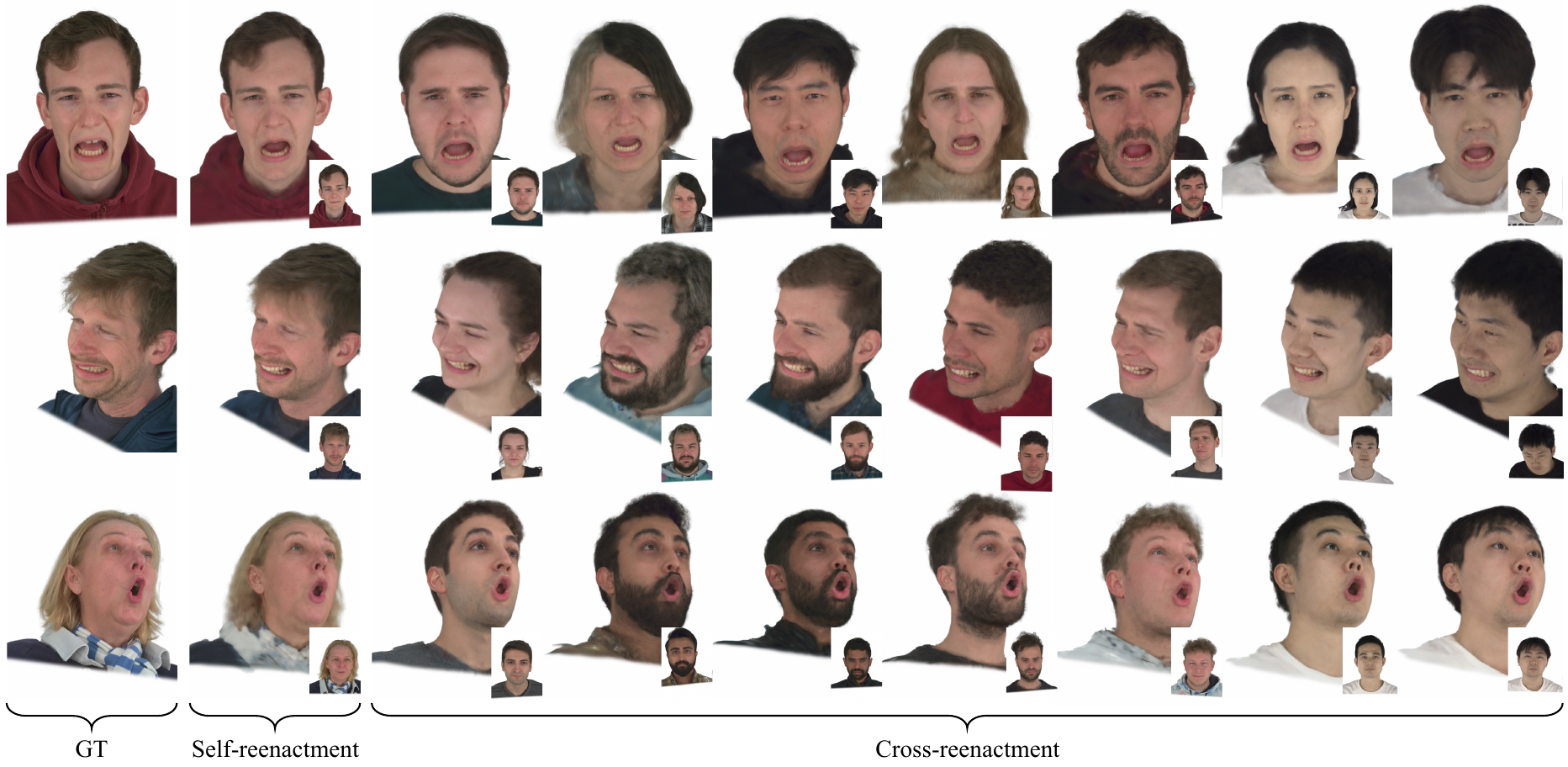}
    \caption{
    Our 3-shot (3-view of the neutral face) results on NeRSemble and our in-house data. From left to right, we show the ground truth, self-reenactment, and cross-reenactment. 
    The lower right of the reenactment results presents one of the three inputs.
    }
    \label{fig:reenactment}
    \vspace{-1em}
\end{figure*}

\begin{figure}[!h]
    \centering
    \includegraphics[width=0.5\textwidth]{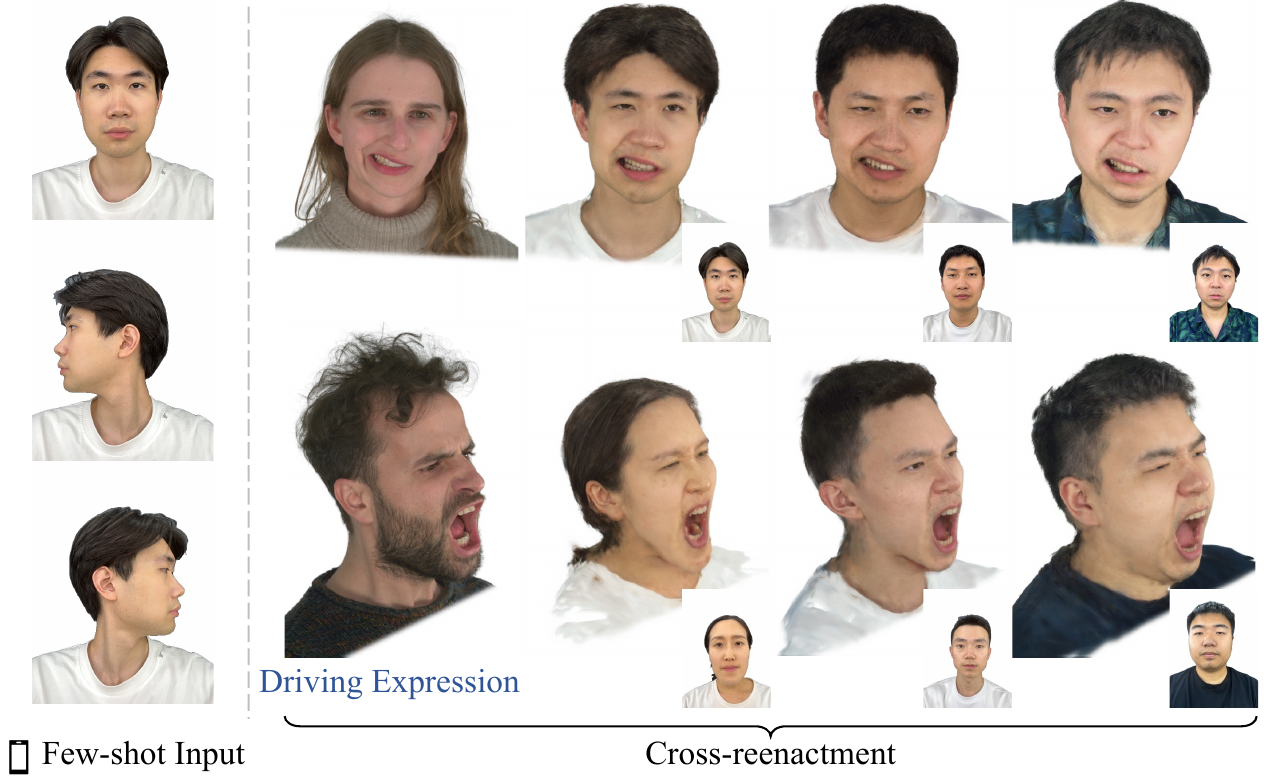}
    \caption{
    Our 3-shot results on in-the-wild data captured by iPad. 
    An example of the input is shown on the left.
    The lower right of the reenactment results presents one of the three inputs.
    }
    \label{fig:phone-captured}
    \vspace{-1em}
\end{figure}

\subsection{Fast Avatar Personalization}
\label{sec:one-shot personalization}

We compare avatar creation with the state-of-the-art on 3 subjects (``074'', ``175'', and ``210') of NeRSemble~\cite{kirschstein2023nersemble}.
To conduct thorough comparisons, we categorize the baselines into methods utilizing: 1) a single image and 2) multiple images.
We analyze the performance of various approaches for both frontal view and all views.
We also present our results using 1-shot and 3-shot inputs.
The quantitative comparisons are listed in \cref{tab:Quantitative comparisons with SOTAs on NeRSemble}. 
We also illustrate qualitative comparisons in \cref{fig:exp-sota-single-view} and \cref{fig:exp-sota-all-view}.

\noindent
\textbf{One-shot personalization.} 
For fair comparisons, our 1-shot results use personalized tracking data from a monocular tracker MICA \cite{zielonka2022towards}, referred to as ``Ours-SV.'' Since we rely on the neck pose from FLAME~\cite{li2017learning}, and MICA lacks this, our performance degrades a lot. Despite this, our approach achieves \custombox{1pt}{gold}{the best} or \custombox{1pt}{silver}{the second best} results across all metrics, demonstrating its robustness.

\noindent
\textbf{Few-shot personalization.} 
\label{sec:3-shot personalization}
Increasing the number of images significantly enhances the authenticity of the head avatar, so we focus on few-shot personalization.
We compare our 3-shot results, using a more precise input mesh from our tracker, with other baseline methods.
Under this setting, our approach achieves \custombox{1pt}{gold}{the best} results across all metrics (Please note that for~\cite{xiang2024flashavatar,qian2024gaussianavatars}, the presence of excessive artifacts makes it impossible to evaluate the ID metrics).
To better show our robustness, we illustrate more of our few-shot results in \cref{fig:reenactment}.
To show the applicability of our approach in real-world scenarios, we also present avatars created from 3 images captured by an iPad in \cref{fig:phone-captured}.
We use these 3 RGB-D images for FLAME fitting.
These results further demonstrate the generalization capability of our approach.

\subsection{Model Analysis}
\label{sec:model-analysis}
We concentrate on few-shot personalization, thus, the experiments primarily examine how our designs affect final few-shot performance.

\begin{table*}[h]
\centering
\makebox[0pt][c]{\parbox{1.05\textwidth}{
\begin{minipage}[h]{0.3\textwidth}
    \centering
    \begin{minipage}[t]{0.9\textwidth}
        \centering
        \scalebox{0.7}{
            \begin{tabular}{lccccccccc}
                \toprule 
                Method & LPIPS$\downarrow$ & PSNR$\uparrow$ & SSIM$\uparrow$ \\

                \midrule
                \multicolumn{3}{l}{\emph{\textcolor{lightgray}{\small{Prior model design:}}}} \\
                \rowcolor{lightblue} Full-model & \textbf{0.140}      & \textbf{22.87} & \textbf{0.854} \\
                w/o Part     & 0.154 & 22.21 & 0.850 \\
                w/o Dynamic  & 0.148 & 22.56 & 0.853 \\
                w/o CNN      & 0.156 & 22.01 & 0.849 \\
                
                \midrule
                \multicolumn{3}{l}{\emph{\textcolor{lightgray}{\small{Few-shot strategy:}}}} \\
                Base (w/o Prior)              & 0.237  &  19.25 & 0.811 \\
                + Inversion                   & 0.171  &  19.67 & 0.829   \\
                \quad  
                + Finetune                    & 0.143  &  22.08 & 0.842  \\
                \rowcolor{lightblue} 
                \quad  \quad 
                + View Reg.                   & \textbf{0.140}      & \textbf{22.87} & \textbf{0.854} \\

                \bottomrule
              \end{tabular}
            
        }
            \caption{Quantitative ablation study. The highlights denote the \colorbox{lightblue}{full-model}.}
            \label{tab:ablation}
    \end{minipage}
\end{minipage}
\hspace{0.5cm}
\begin{minipage}[h]{0.7\textwidth}
	\centering
	\includegraphics[width=\textwidth]{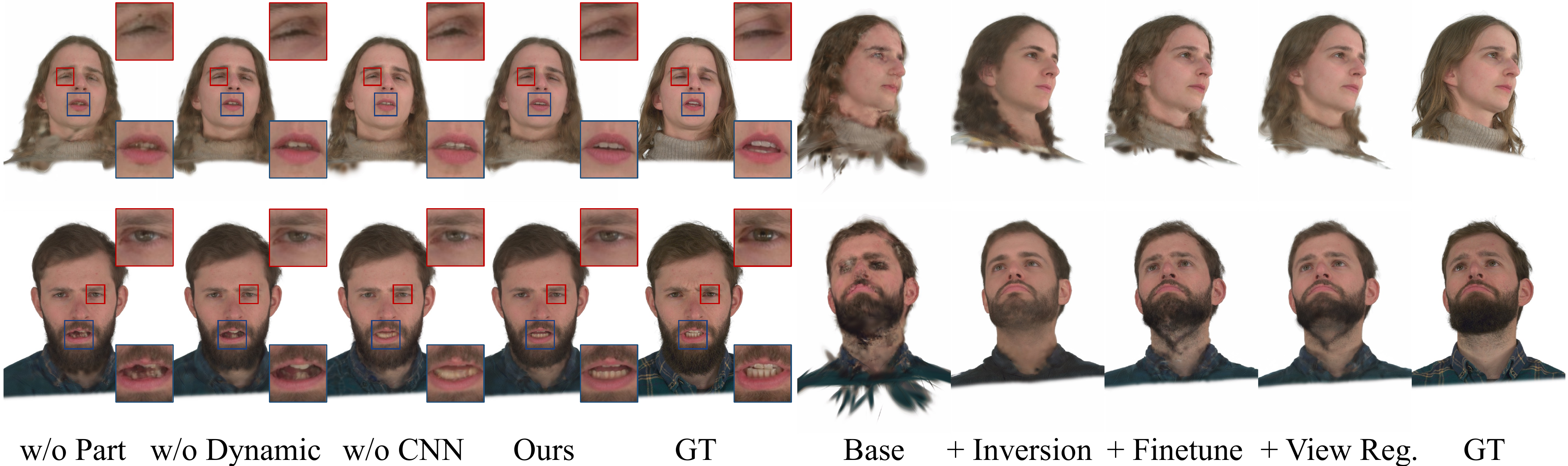}
        \captionof{figure}{Qualitative ablation study. Please zoom in for more details.}
	\label{fig:ablation}
\end{minipage}
}}
\vspace{-8px}
\end{table*}

\begin{figure*}[h]
    \centering
    \begin{subfigure}[b]{0.33\textwidth}
        \centering
        \includegraphics[width=\textwidth]{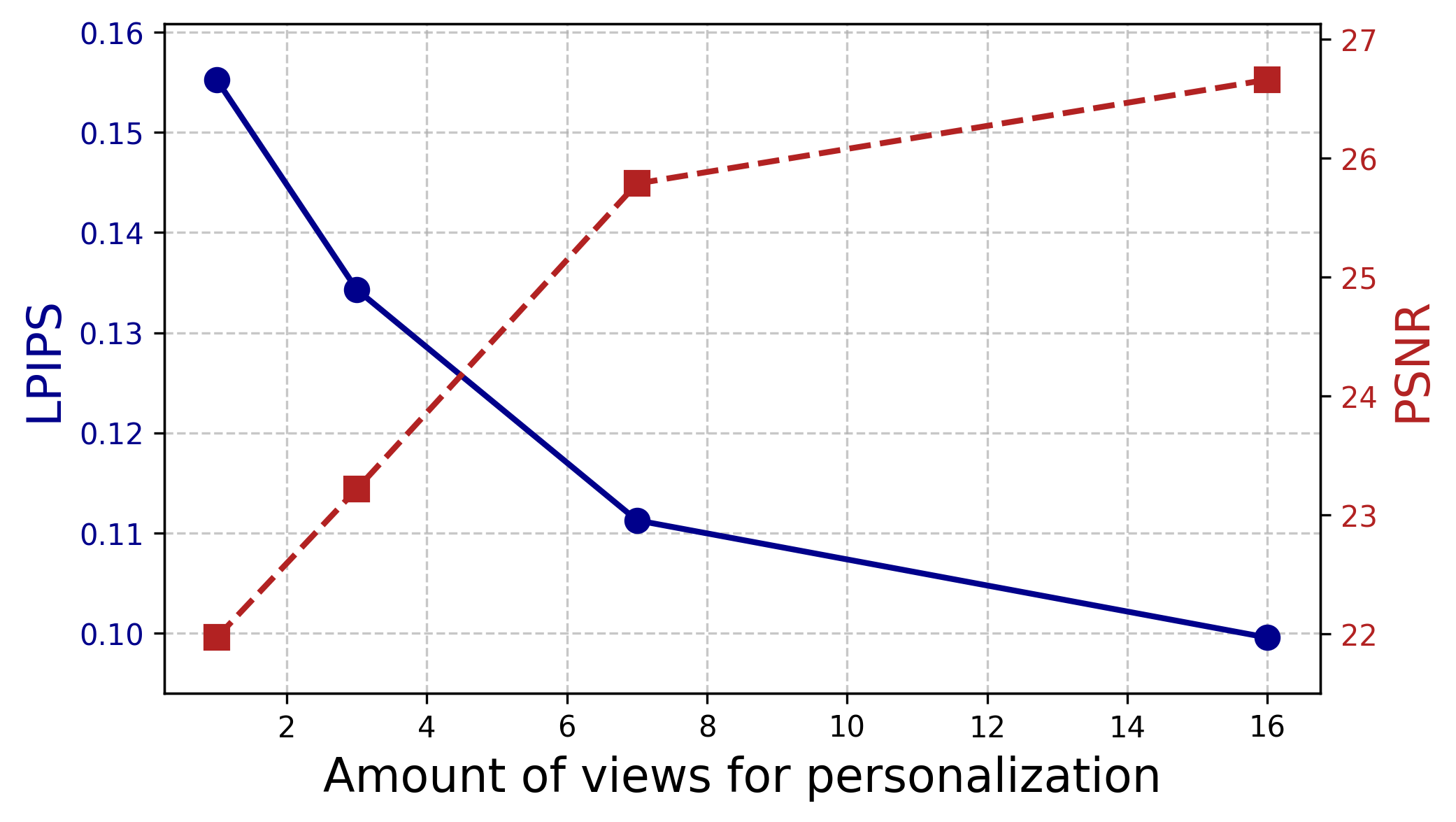}
        \caption{Quality w.r.t \#personalizing data.}
        \label{fig:sub-figure1}
    \end{subfigure}
    \hfill
    \begin{subfigure}[b]{0.33\textwidth}
        \centering
        \includegraphics[width=\textwidth]{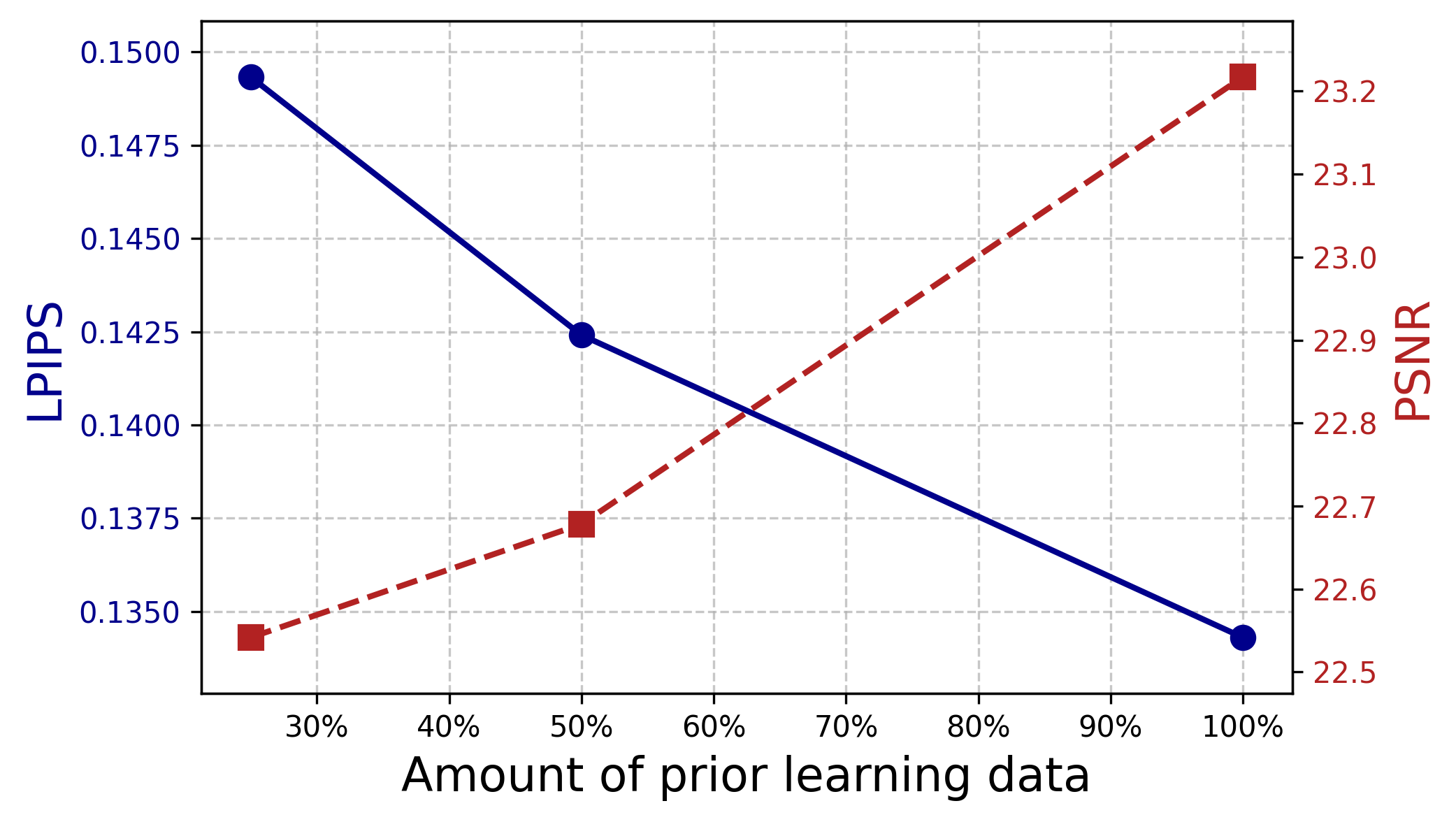}
        \caption{Quality w.r.t \#prior learning data.}
        \label{fig:sub-figure2}
    \end{subfigure}
    \hfill
    \begin{subfigure}[b]{0.33\textwidth}
        \centering
        \includegraphics[width=\textwidth]{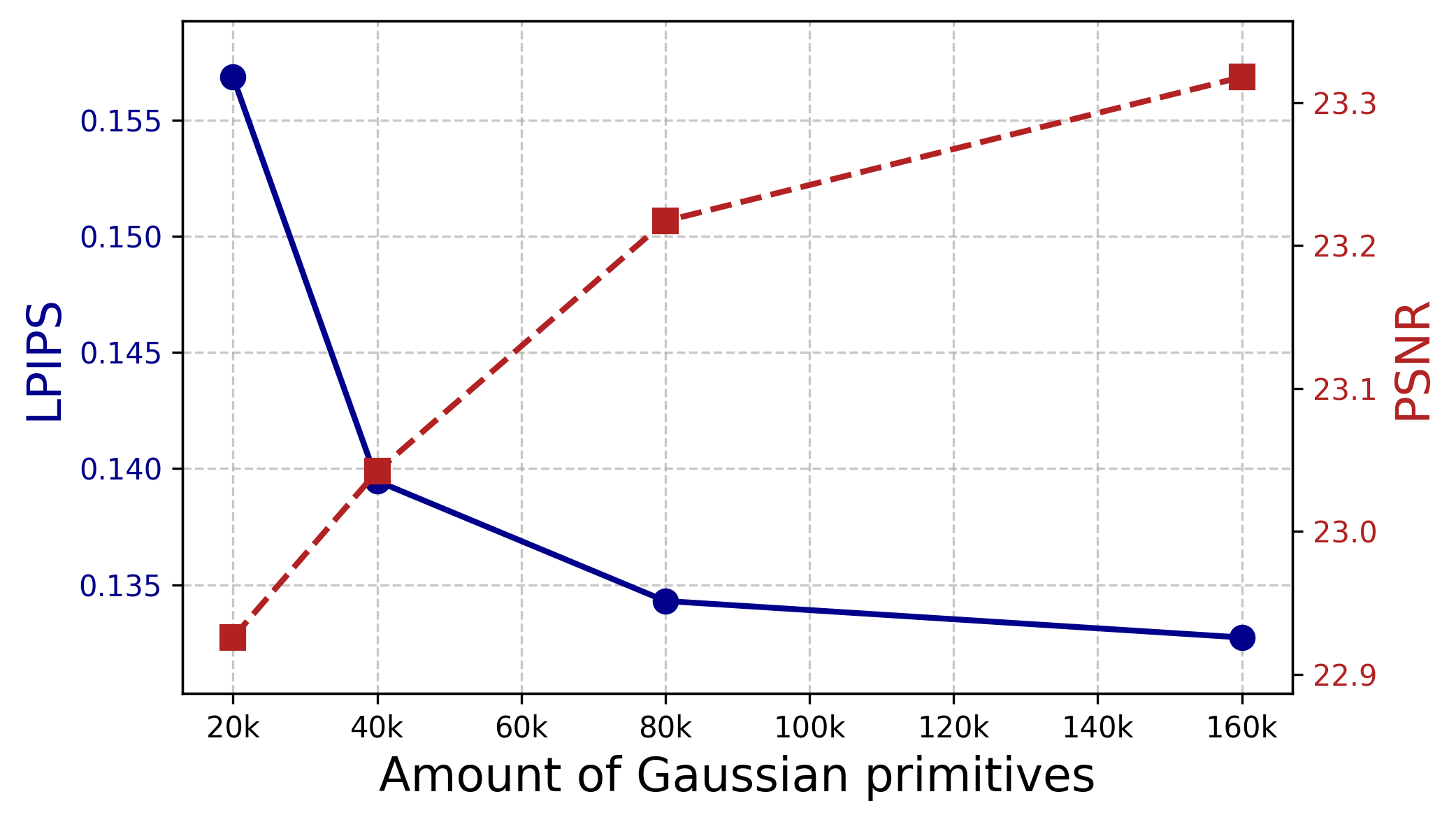}
        \caption{Quality w.r.t \#primitives.}
        \label{fig:sub-figure3}
        
    \end{subfigure}
    
    \label{fig:three_figures}
    \vspace{-1em}
\end{figure*}

\noindent
\textbf{Ablation.}
We use subjects ``256'' and ``270'' for conducting the ablation study.
The ablation is divided into two parts, with the first part to validate our model designs and the second part to justify our few-shot strategies.
\cref{tab:ablation} presents the quantitative results for the testing sequence.
Moreover, we illustrate the qualitative results in \cref{fig:ablation}.

For model design ablations, we remove one component at a time to demonstrate their effectiveness.
Excluding part-based modeling (``w/o Part'') causes a significant performance drop, as it hampers effective prior learning and part-based few-shot strategies.
Not modeling dynamic information (``w/o Dynamic'') also degrades performance, making it challenging to capture details in highly dynamic regions (\eg, the mouth).
Lastly, using CNN for refinement reduces artifacts and enhances detail realism (``w/o CNN'').

3DGS-based methods have strong input-fitting capabilities, leading to overfitting with limited training data and poor generalization to novel views and expressions (``Base''). 
Thus, 3DGS-based methods heavily rely on prior knowledge for few-shot personalization. 
Using our prior model for inversion (``+ Inversion'') produces avatars similar to the input data with robust animation.
Additional fine-tuning (``+ Finetune'') enhances realism with personalized details. 
View regularization (``+ View Reg.'') helps reduce novel-view artifacts.

\noindent
\textbf{Quality w.r.t quantity of personalizing data.} 
HeadGAP supports various numbers of inputs, allowing us to analyze few-shot performance with different data amounts. The analysis has two parts:
1.~Quantitative results with different numbers of views of the neutral face (\cref{fig:sub-figure1}) show that performance improves as the number of views increases.
2.~Performance with 8 additional images of different frontal view expressions (\cref{fig:more-input}) demonstrates that more inputs help the model capture personalized dynamic details.

{
\setlength{\abovecaptionskip}{5pt plus 3pt minus 2pt} 
\begin{figure}[h]
    \centering
    \includegraphics[width=1.\columnwidth]{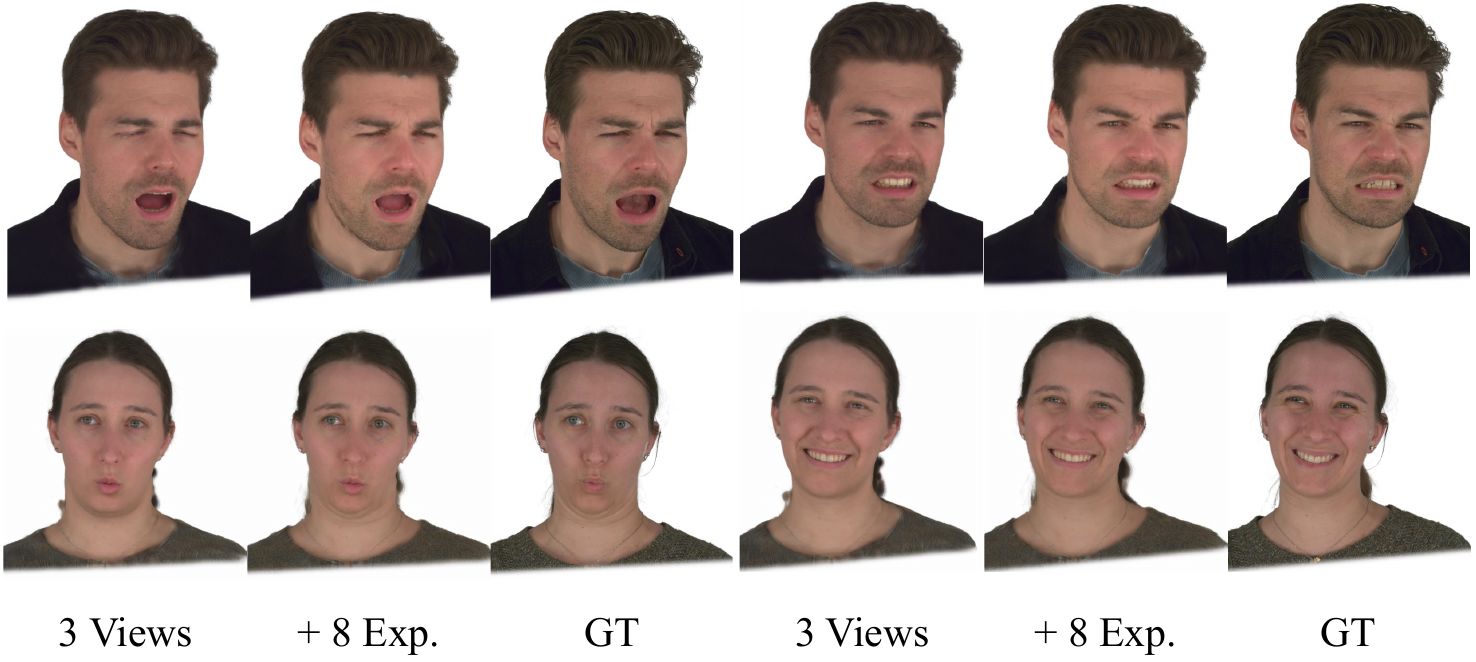}
    \caption{
    Comparisons between using 3-view data and 3-view with 8 additional expressions' data of the frontal view.
    }
    \label{fig:more-input}
\end{figure}
}

\noindent
\textbf{Quality w.r.t quantity of prior learning data.}
The large-scale prior learning data is the core of the high-fidelity few-shot personalization. 
Therefore, we conduct analysis to show our performance with varying amounts (IDs) of data for prior learning.
As depicted in \cref{fig:sub-figure2}, more prior learning data makes the final avatar more realistic.
Moreover, we do not observe significant saturation, indicating our model can benefit from more available 3D data.
It is worth noting that our model is robust for different amounts of prior learning data due to our complete pipeline.

\noindent
\textbf{Quality w.r.t quantity of primitives.}
We also evaluate our performance with varying numbers of Gaussian primitives. 
\cref{fig:sub-figure3} proves our method can create more realistic avatars with more primitives. 
To balance the fidelity and efficiency, we utilize 80k primitives for our default model. 
We do not focus on finding the optimal method to control the number of primitives.
We believe our approach could also benefit from other adaptive density control approaches~\cite{kerbl20233d}.

\noindent
\textbf{Network comparison.}
GAPNet is capable of adapting to different numbers of IDs for training. To demonstrate the network capability, we compare its performance for a single person against~\cite{qian2024gaussianavatars}.
GAPNet achieves better performance in single-person modeling, with an LPIPS/PSNR of 0.091/25.48 compared to GaussianAvatars' 0.119/25.32.

\vspace{-.4em}
\section{Conclusion}
\label{sec:conclusion}
In this paper, we present a novel approach for creating high-fidelity 3D head avatars with few-shot images. 
We first learn 3D Gaussian priors from large-scale 3D head data, then create avatars of the novel identities with the aid of the priors.
To facilitate the learning of powerful and generalizable priors, we develop GAPNet which can exploit 3D part-based dynamic head priors and 2D structured head priors for creating high-fidelity avatars with robust animations.
The comprehensive experiments justify our designs and superiority.
We also showcase our robustness by creating avatars of plentiful identities from the public dataset and images captured by consumer-grade devices.

\maketitlesupplementary

\renewcommand{\thesection}{\Alph{section}} 
\renewcommand{\thetable}{\Alph{table}}  
\renewcommand{\thefigure}{\Alph{figure}}

\setcounter{section}{0}
\setcounter{figure}{0}
\setcounter{table}{0}

\section{Implementation Details}
\noindent\textbf{Dataset.}
We partition the data into training and testing sets, comprising 119 and 45 subjects, respectively.
Of these subjects, the data of 11 training subjects and 3 testing subjects are provided by \cite{qian2024gaussianavatars} and others are processed by our FLAME tracking algorithm.
For more information about the dataset, we highly encourage the reader to refer to the paper of NeRSemble~\cite{kirschstein2023nersemble} for further details.

\begin{figure}[ht]
    \centering
    \includegraphics[width=1.\columnwidth]{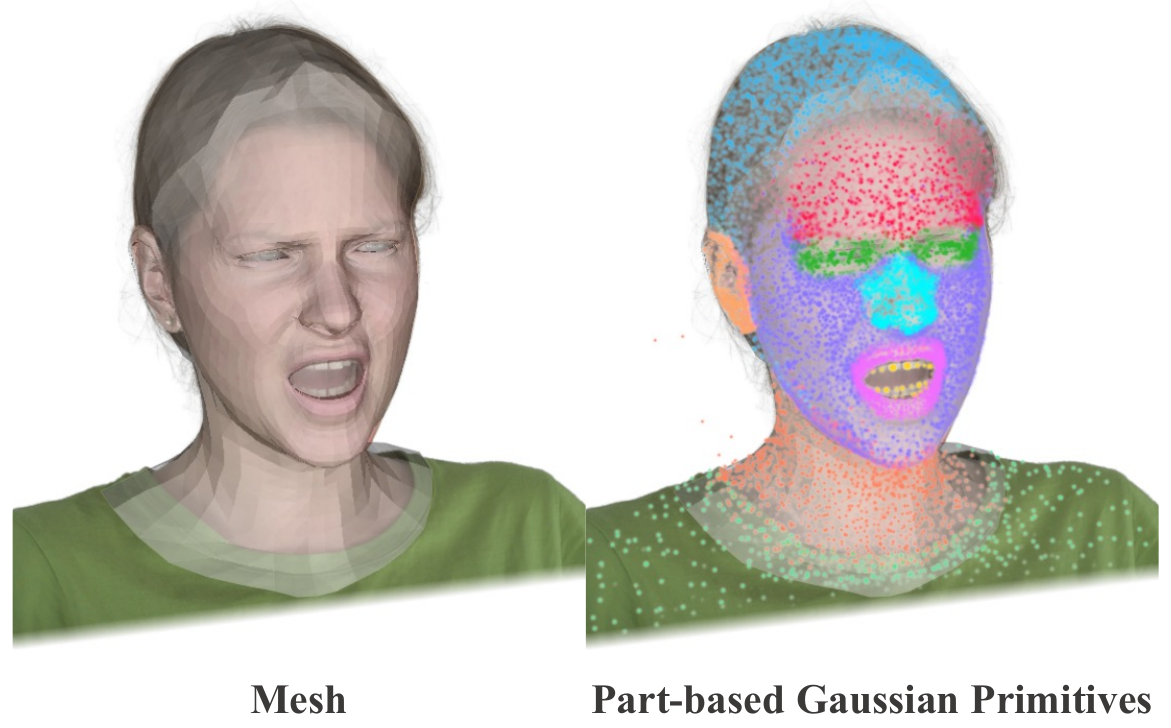}
    \caption{
    Illustration of the FLAME mesh (left) and semantic part of our Gaussian primitives bound on the mesh (right). Different point colors represent different parts.
    }
    \label{fig:supp-part}
\end{figure}

\noindent\textbf{Model Detail.}
We divide the Gaussian primitives into $p=11$ parts, including 1) ``forehead'', 2) ``nose'', 3) ``eye'', 4) ``teeth'', 5) ``lip'', 6) ``ear'', 7) ``hair'', 8) ``boundary'', 9) ``neck'', 10) ``other face region'', and 11) ``other''.
The part for the primitives is determined by the face masks provided by FLAME \cite{li2017learning}. 
The illustration of Gaussian primitives with different parts is shown in \cref{fig:supp-part}. 
The primitive number is set to $n=83,651$ by initializing from a UV map with a resolution of $300\times 300$.
The feature dimensions of identity-shared point encoding $\mathbf{f}$, identity code $\mathbf{z}$, and point appearance feature $\mathbf{h}$ are set to $c_{1} = 48$, $c_{2} = 128$, and $c_{3} = 34$ respectively.
All the MLPs $f^{\mathcal{M}}$ consist of $4$ layers. 
Meanwhile, the CNN $f^{\mathcal{C}}$ contains $6$ layers. 
The identity codebook $\mathbf{z}$ is initialized with zero.

\noindent\textbf{Training Detail.}
We adopt Adam~\cite{kingma2014adam} optimizer for the model training.
For prior learning, we utilize $k=119$ identities and set the batch size to $32$.
For all the parameters, the learning rate begins at $1e^{-3}$ and decreases with the cosine scheduler.
The prior model is trained on $8$ A100 GPUs for 100K steps, which takes around 2 days.
The loss weights $\lambda_{m}$, $\lambda_{l1}$, $\lambda_{ssim}$, $\lambda_{lpips}$, $\lambda_{\alpha}$, $\lambda_{s}$, $\lambda_{\mu}$, and $\lambda_{arap}$ are set to $10$, $0.8$, $0.2$, $0.4$, $1$, $1$, $0.01$, and $1$ respectively.
For few-shot personalization, we set the batch size to $1$. 
We set the learning rate of the identity-shared point encoding $\mathbf{f}$ to $1e^{-3}$ and other parameters' to $1e^{-5}$.
Unless otherwise stated, we take $500$ steps for inversion and $500$ steps for fine-tuning, which uses about $5$ minutes in total with an A100 GPU. 
For view regularization, we generate $m=16$ reference views similar to the camera setups of the NeRSemble dataset.
The loss weights $\lambda_{ref}$ is set to $0.01$.
For 3-shot novel identities' personalization on NeRSemble, we utilize cameras with id ``0'', ``8'', and ``15''. 
For our captured data, we select viewpoints similar to those of NeRSemble.

\noindent\textbf{About adaptive density control.}
To allow full control of primitive numbers, we do not utilize adaptive density control, opacity reset, and point pruning as GaussianAvatars~\cite{qian2024gaussianavatars}.

\section{Experiment Results}

\subsection{Network Comparison}
\begin{figure*}[b!]
    \centering
    \includegraphics[width=2.0\columnwidth]{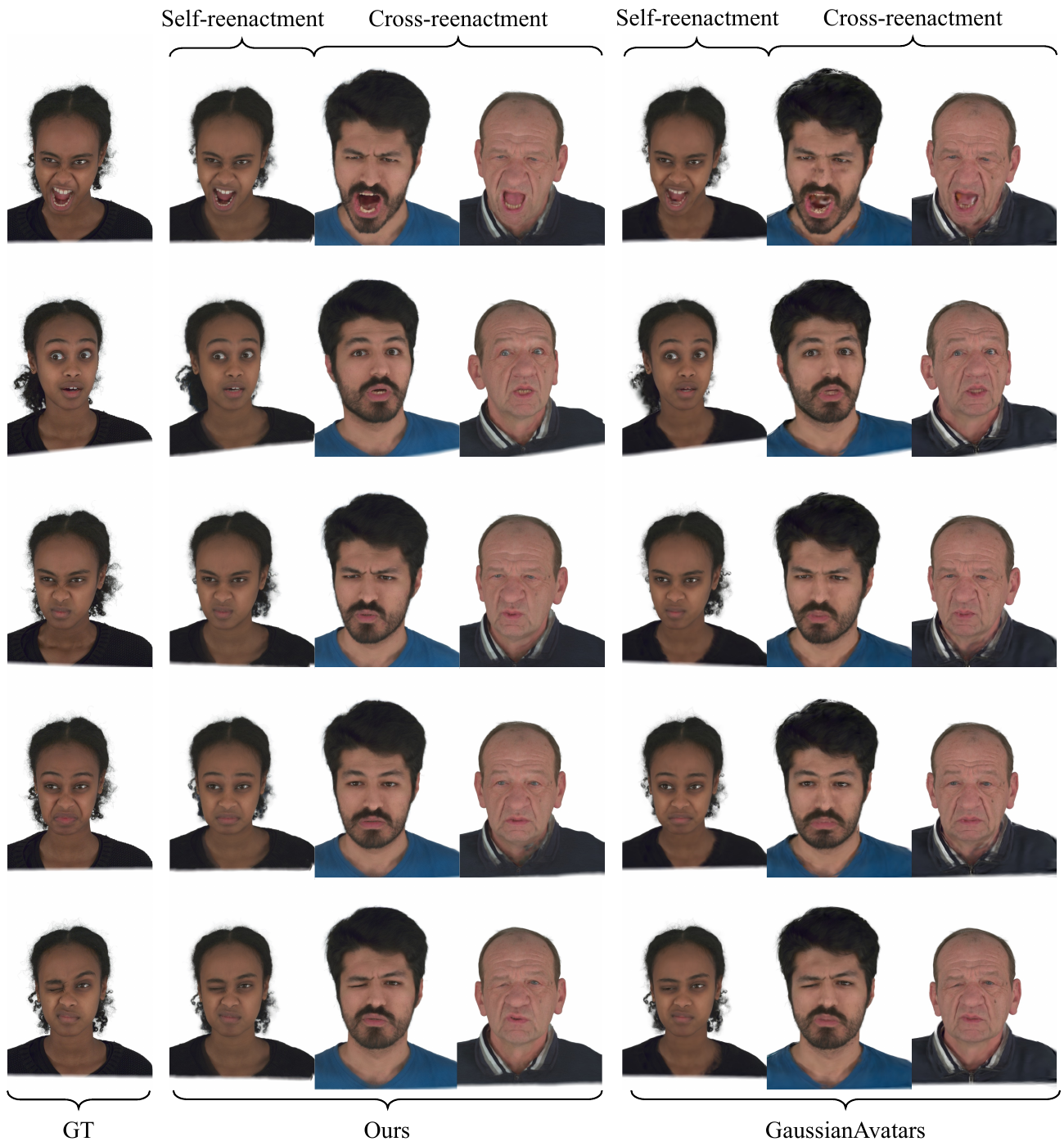}
    \caption{
    Self- and cross-reenactment comparisons between our method and GaussianAvatars for single-subject modeling.}
    \label{fig:supp-network-comp}
\end{figure*}

GAPNet is capable of adapting to different numbers of IDs for training.
To demonstrate the network capability, we compare its performance for a single person against GaussianAvatars~\cite{qian2024gaussianavatars}.
For a fair comparison of the network, we utilize the same adaptive density control approaches as~\cite{qian2024gaussianavatars}.
We also use the full training data of each single subject, similar to~\cite{qian2024gaussianavatars}.
The mean quantitative results over subject ``074'', ``175'', and ``210'' are shown in \cref{tab:single-person}.
GAPNet obtains better results in all metrics.
We further illustrate the qualitative comparisons in \cref{fig:supp-network-comp}. 
Our model is capable of fitting the dynamic details of the training subject well, as shown in self-reenactment results. 
Moreover, our cross-reenactment performance is significantly more robust than GaussianAvatars. 
The robust animations further prove our model design is quite suitable for learning generalizable priors across different subjects.

\begin{table}[h]
  \centering
  \begin{tabular}{lccccccccc}
    \toprule 
    Method & LPIPS$\downarrow$ & PSNR$\uparrow$ & SSIM$\uparrow$ \\
    \midrule
    GaussianAvatars & 0.120  & 25.21 & 0.911  \\
    Ours            & \textbf{0.091}  & \textbf{25.48} & \textbf{0.912}  \\
    \bottomrule
  \end{tabular}
  \caption{Comparisons for single avatar creations.}
  \label{tab:single-person}
\end{table}

\subsection{Prior Learning Results}
We show our prior learning results of the 119 identities in \cref{fig:prior-learning-ids}.
The visualized results show that our GAPNet can learn the appearance characteristics of different identities.

\subsection{More Qualitative Results}
\label{sec:supp-more-results}
\begin{figure*}[h!]
    \centering
    \includegraphics[width=\textwidth]{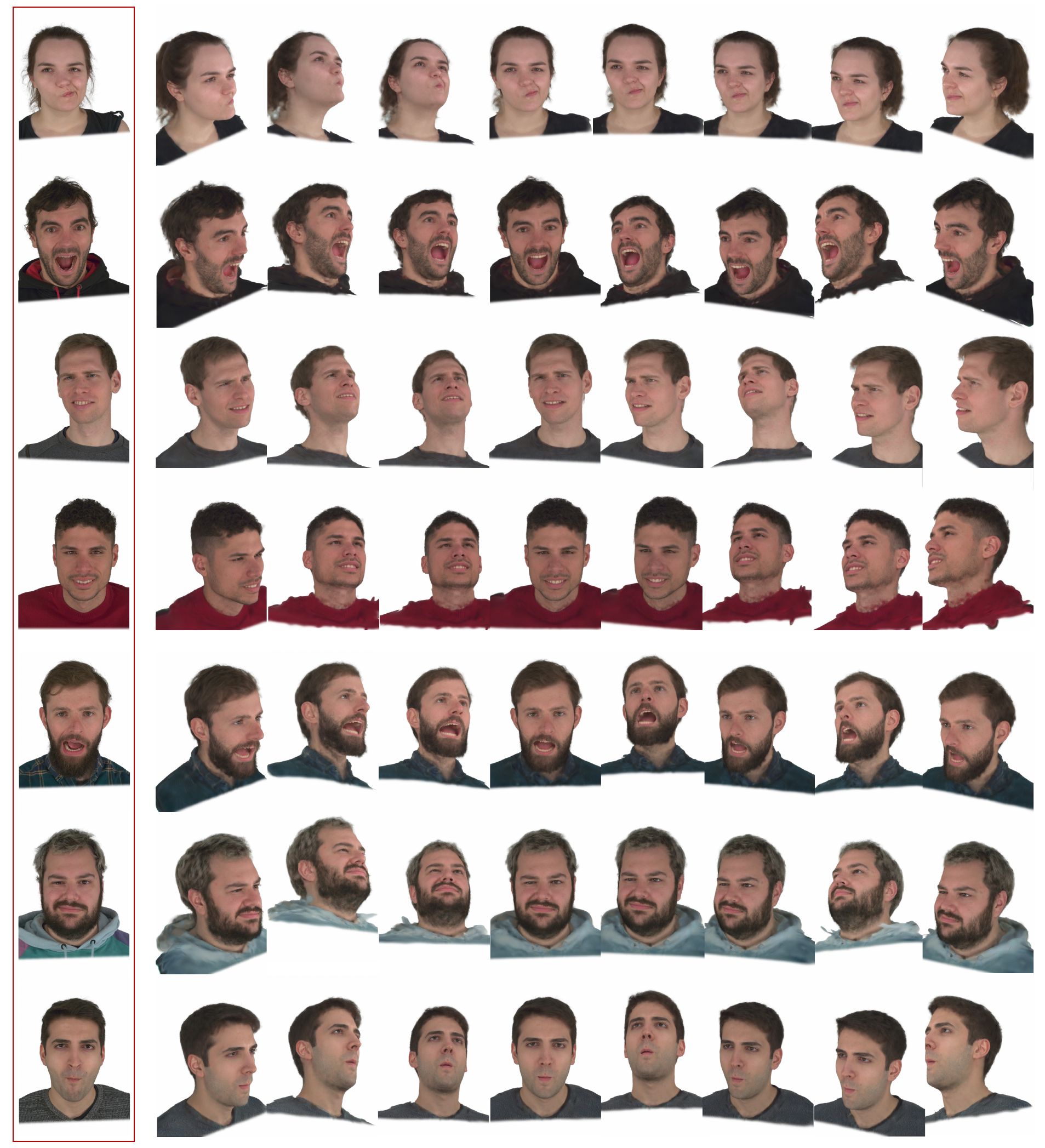}
    \caption{
    Self-reenactment results. The images inside the red box are the driving expressions. We showcase the renderings from different viewpoints.
    }
    \label{fig:supp-more-self}
\end{figure*}

\begin{figure*}[h!]
    \centering
    \includegraphics[width=\textwidth]{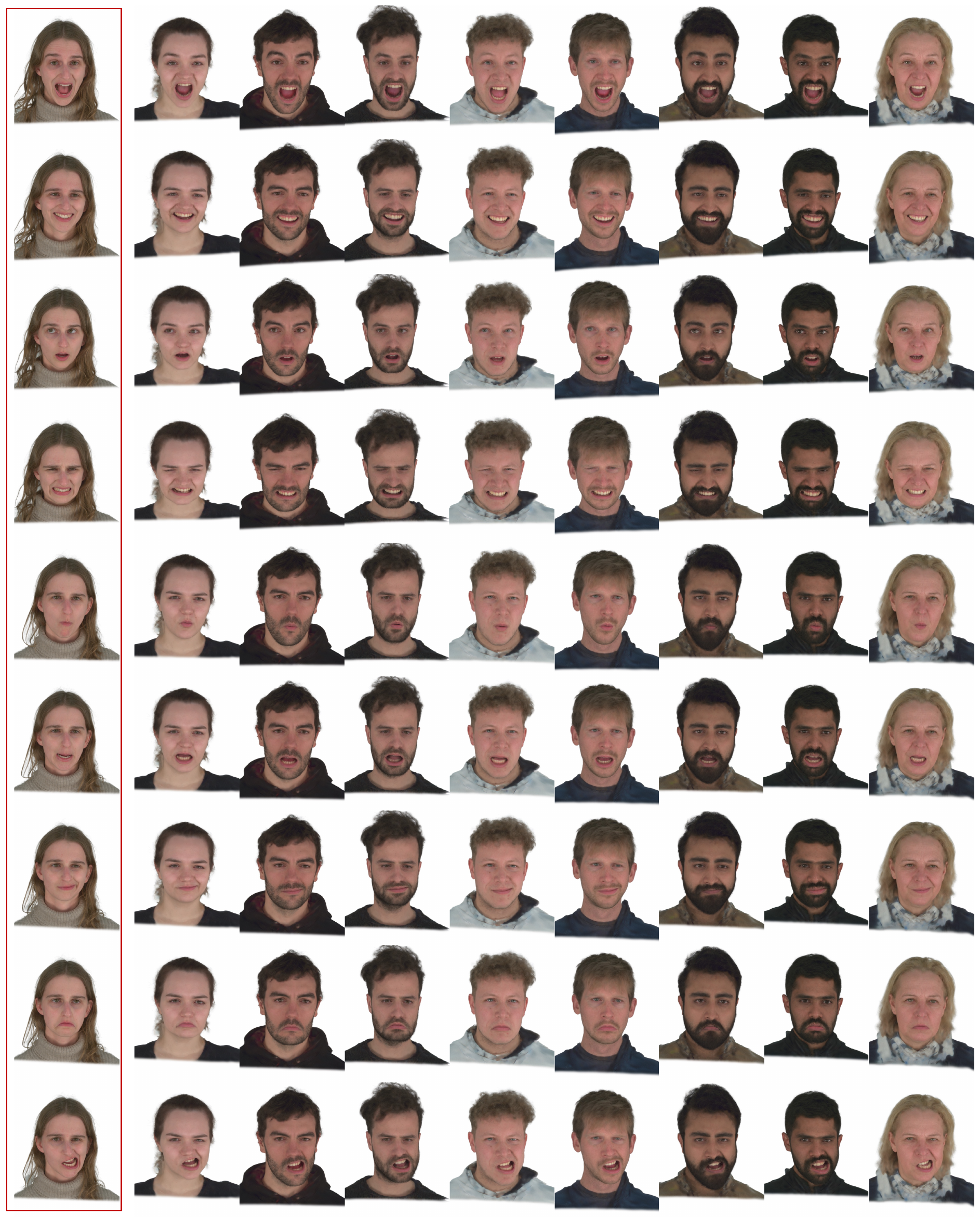}
    \caption{
    Cross-reenactment results. The images inside the red box are the driving expressions.
    }
    \label{fig:supp-more-cross}
\end{figure*}

\begin{figure*}[h!]
    \centering
    \includegraphics[width=\textwidth]{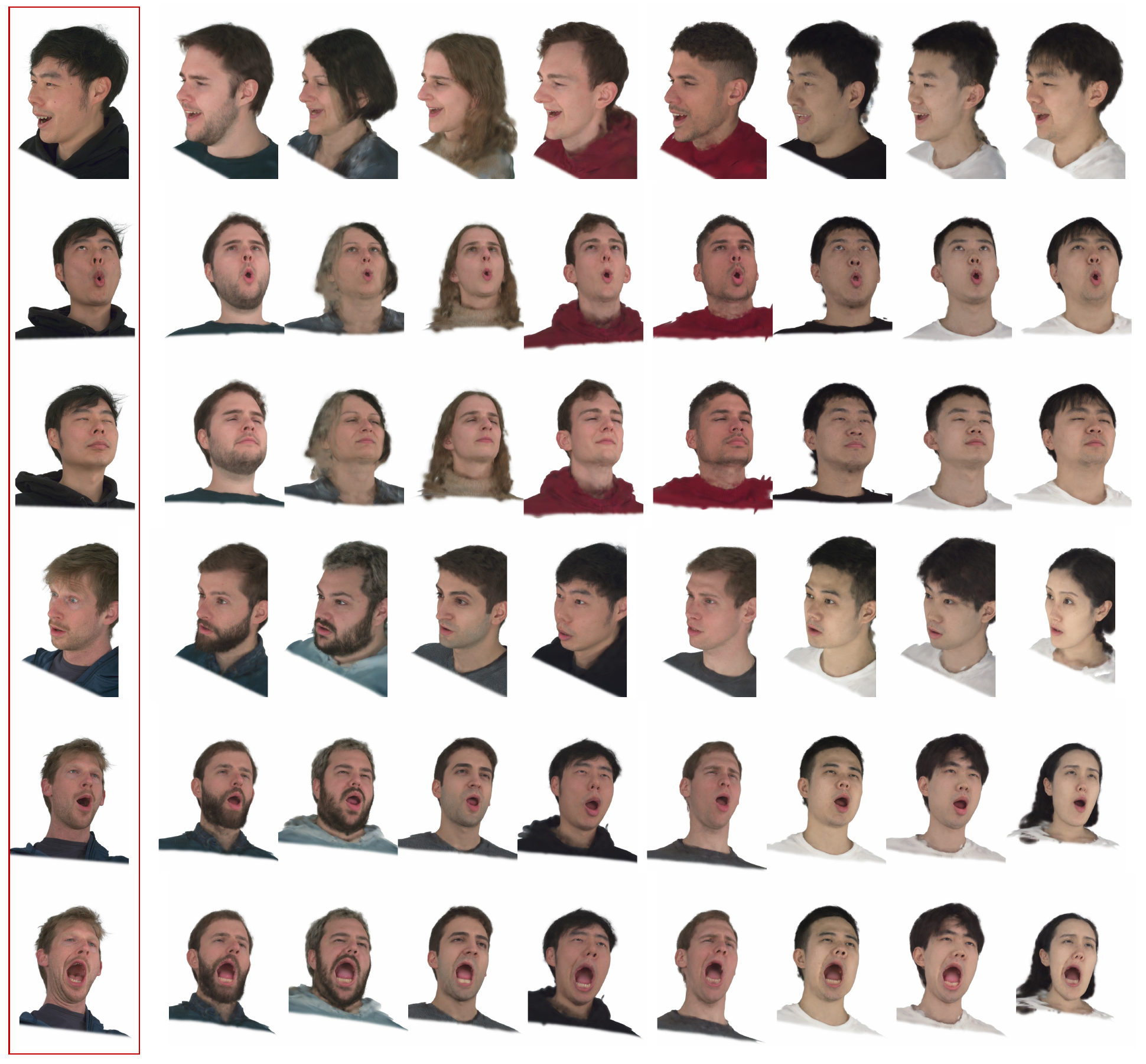}
    \caption{
    Cross-reenactment results. The images inside the red box are the driving expressions.
    }
    \label{fig:supp-more-cross-2}
\end{figure*}

We present additional qualitative experimental results in \cref{fig:supp-more-self}, \cref{fig:supp-more-cross}, and \cref{fig:supp-more-cross-2}.
All subjects are novel IDs and were not seen during the training process.

\cref{fig:supp-more-self} shows self-reenactment results with novel-view renderings for different identities. \cref{fig:supp-more-cross} and \cref{fig:supp-more-cross-2} present cross-reenactment results from frontal and side views, respectively, demonstrating stable animations.

The results indicate that our model effectively generalizes to data that differs from the NeRSemble dataset. Furthermore, it achieves consistent few-shot performance across diverse ethnicities and genders, thereby further reinforcing its capacity for effective generalization.

\subsection{Head Avatar Editing}
\begin{figure*}[b!]
    \centering
    \includegraphics[width=\textwidth]{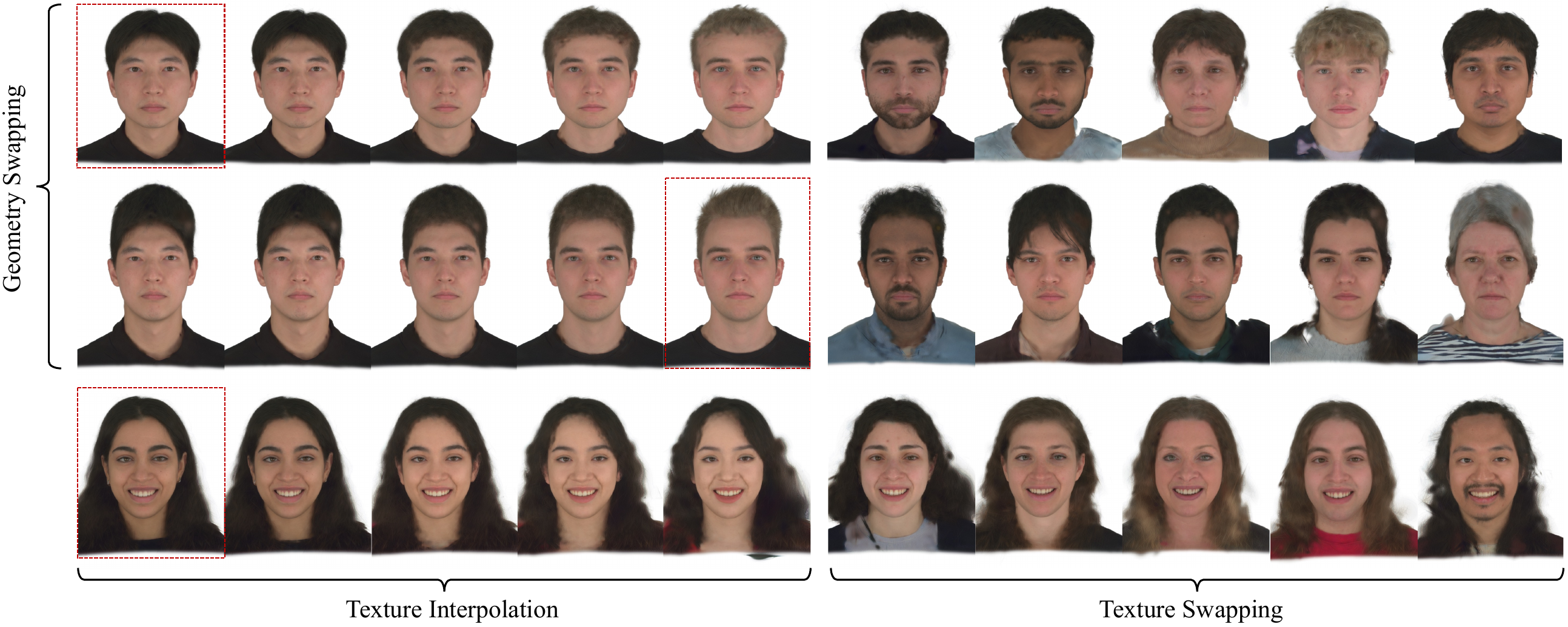}
    \caption{
    Illustration of the GAPNet's 1) texture interpolation, 2) texture swapping, and 3) geometry swapping. The results on the same row are using the same head geometry. 
    The identities inside red boxes use the paired texture and FLAME mesh.
    }
    \label{fig:interpolation}
\end{figure*}

Since our representation models textures using 3D Gaussian Splatting upon the base FLAME mesh, we can perform 1) texture interpolation between different identities using the same FLAME mesh, 2) texture swapping using the same FLAME mesh, and 3) geometry editing by swapping the FLAME mesh.
The results are shown in \cref{fig:interpolation}. 

\subsection{More In-the-wild Results}
\begin{figure*}[b!]
    \centering
    \includegraphics[width=\textwidth]{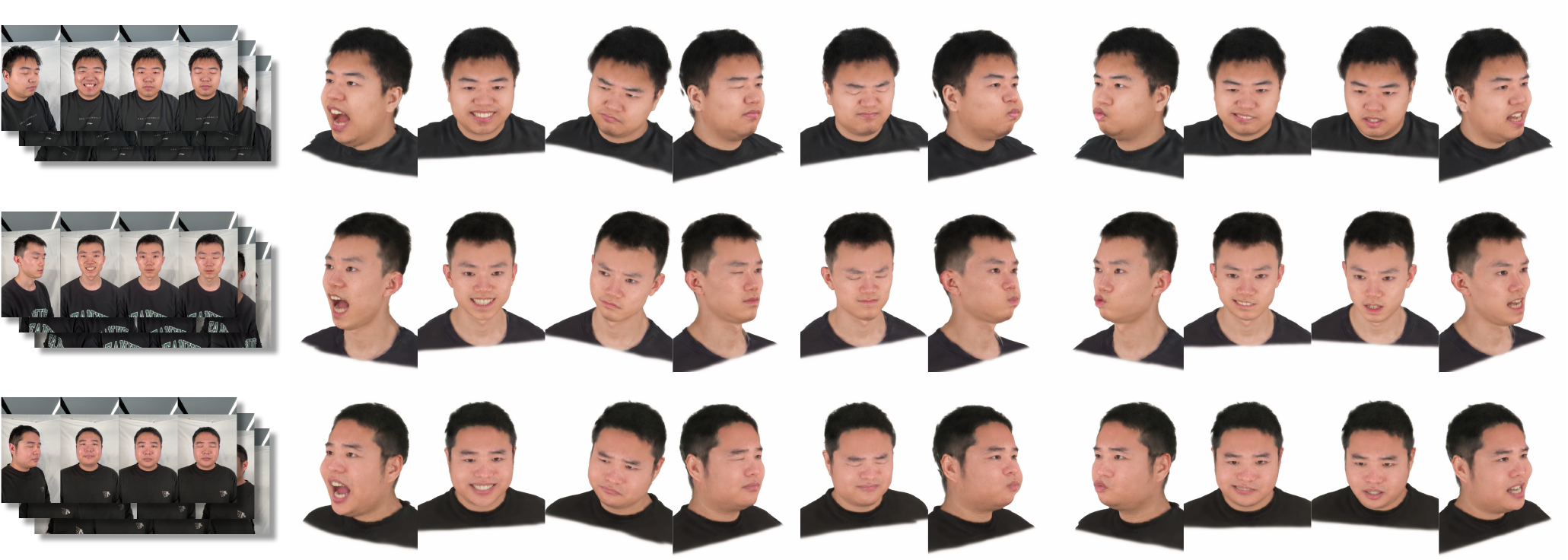}
    \caption{
    Qualitative results of 3D animatable head avatars generated from few-shot in-the-wild images and driven by the same facial expression sequence.
    }
    \label{fig:itw}
\end{figure*}

\begin{figure*}[b!]
    \centering
    \includegraphics[width=\textwidth]{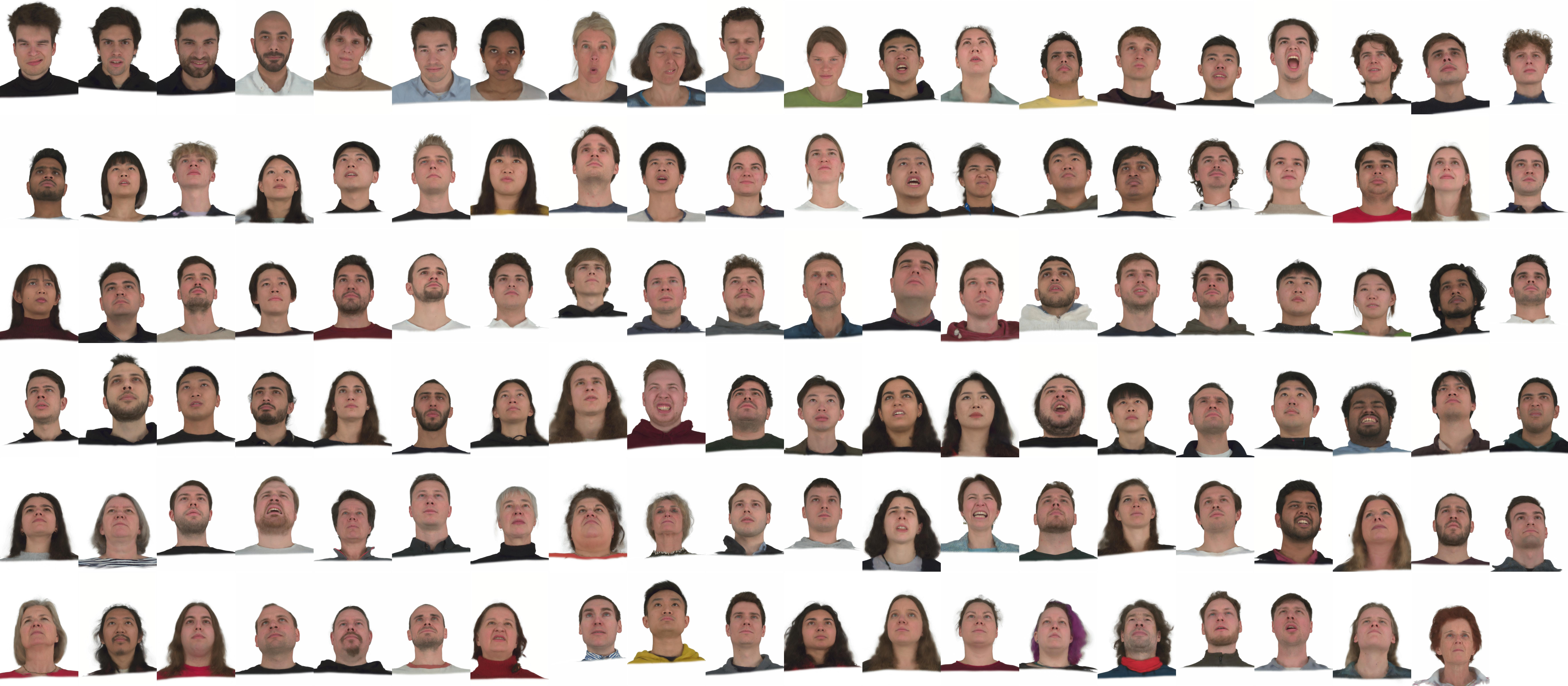}
    \caption{
    The rendered results of the 119 identities used for prior learning.
    }
    \label{fig:prior-learning-ids}
\end{figure*}

In this section, we present additional results on in-the-wild images. All result IDs are out-of-distribution samples beyond the NeRSemble~\cite{kirschstein2023nersemble} dataset. Specifically, we capture monocular video data of each identity performing various expressions and select 12 images for avatar personalization. As shown in ~\cref{fig:itw}, we present the cross-reenactment driving results when providing the same facial expression motion sequence. The results demonstrate that our method exhibits strong few-shot generalization capability even in in-the-wild settings.

\section{Further Discussions on Baselines}
We compare multiple baseline approaches for one-shot and few-shot personalization based on the number of input images in the main text.
In this section, we further elucidate the details of the experimental comparisons.

\subsection{Baseline taxonomy}
We categorize the baselines into two types based on whether they involve a process of learning priors.

\textbf{Type-\RNum{1}} includes:
ROME~\cite{khakhulin2022realistic}, GOHA~\cite{li2024generalizable}, VOODOO 3D~\cite{tran2024voodoo}, HiDe-NeRF~\cite{li2023one}, Portrait4Dv1~\cite{deng2024portrait4d}, Portrait4Dv2~\cite{deng2024portrait4dv2}, GPAvatar~\cite{chu2024gpavatar} and DiffusionRig~\cite{ding2023diffusionrig}.

\textbf{Type-\RNum{2}} includes: FlashAvatar~\cite{xiang2024flashavatar}, GaussianAvatars~\cite{qian2024gaussianavatars} and NHA~\cite{grassal2022neural}

\subsection{Comparison with single-view baselines}
In one-shot personalization experiments, when driving novel expressions, tri-plane representation-based volume rendering methods~\cite{li2024generalizable,tran2024voodoo,li2023one,deng2024portrait4d,deng2024portrait4dv2,chu2024gpavatar} require the driving image as input. This might result in appearance leakage (\eg, dynamic details of new expressions). In contrast, our method uses only the tracking mesh of the driving image and models dynamic details through prior learning.

\subsection{Comparison with GS-based methods}
We show the comparison with GaussianAvatars~\cite{qian2024gaussianavatars} and FlashAvatar~\cite{xiang2024flashavatar} in \cref{fig:supp-gaussian-views}. Although they do not focus on few-shot input like ours, we include comparisons because we all use 3D Gaussian Splatting as a representation. 
We observe that they require a substantial overlap of input views or monocular videos with human heads rotated to different orientations.

In few-shot personalization experiments, as shown in the~\cref{fig:supp-gaussian-views}, all per-subject optimization Gaussian Splatting-based baselines lack prior information and require individual training for each person. It can be observed that all baseline methods tend to overfit the training views and fail to extrapolate to unseen views. This qualitative comparison demonstrates the effectiveness and necessity of constructing priors for Gaussian Splatting.
Due to the noticeable artifacts, FlashAvatar$^\dagger$, GaussianAvatars$^\ddagger$, and GaussianAvatars$^\blacklozenge$ are infeasible for calculating meaningful ID similarity metrics.
Therefore, we did not report their corresponding metrics in Tab.1 of the main text.

\section{Limitations and Future Works}
While our method can quickly construct personalized, high-fidelity, and realistic human head avatars, it still has the following issues:
(1) In cases where the subject wears glasses or has noticeable facial accessories, the avatar construction may exhibit artifacts (as depicted in \cref{fig:supp-failure}).
A reason for this incapability is that our prior learning phase does not incorporate such samples for training. 
Including the corresponding data for training can potentially resolve this problem.
(2) The adoption of CNNs for refinement in screen space may result in view-dependent overfitting, which can induce flickering among different viewpoints and lead to quality degradation for certain unseen views during training. Therefore, exploring more consistent refinement techniques in 3D space presents a promising avenue for further investigation.
(3) Our method does not focus on modeling the subject's clothing and hair. We believe that combining methods such as \cite{luo2024gaussianhair} to model hair or clothing separately is a promising research direction.
(4) Additionally, lighting variation is important for the realism of head avatars. Currently, we only consider uniform lighting. We believe that integrating relighting into the Gaussian Splatting is also a promising research direction for head avatars.

\clearpage

\clearpage

\begin{figure*}[h!]
    \centering
    \includegraphics[width=\textwidth]{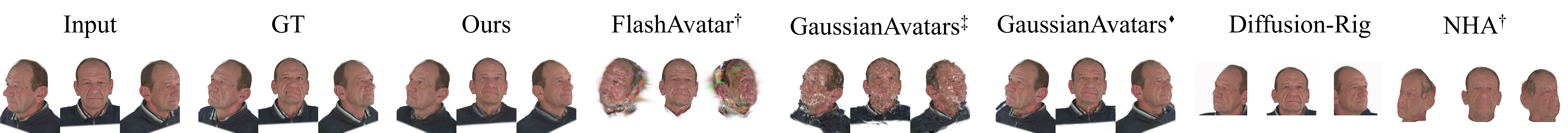}
    \caption{
    More qualitative experiments on other subjects using 3-shot inputs compared to state-of-the-art methods.
    }
    \label{fig:supp-sota-all-view}
    \vspace{2em}
\end{figure*}

\begin{figure*}[b]
    \centering
    \includegraphics[width=\textwidth]{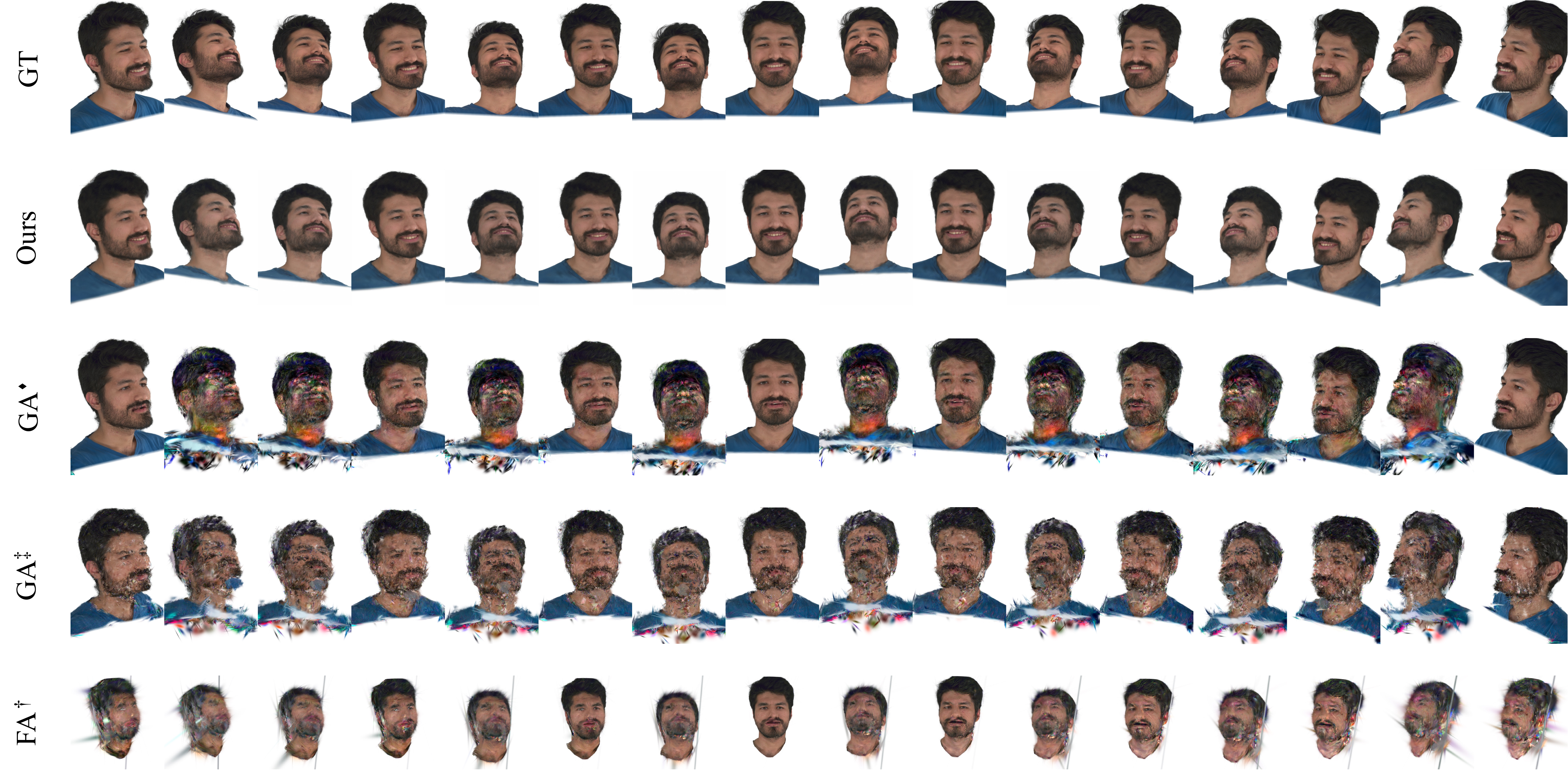}
    \caption{Qualitative comparison results. We compare the rendering results from different views using our 3-shot input avatars with the Gaussian Splatting-based baseline methods.}
    \label{fig:supp-gaussian-views}
\end{figure*}

\begin{figure*}[b!]
    \centering
    \includegraphics[width=2.0\columnwidth]{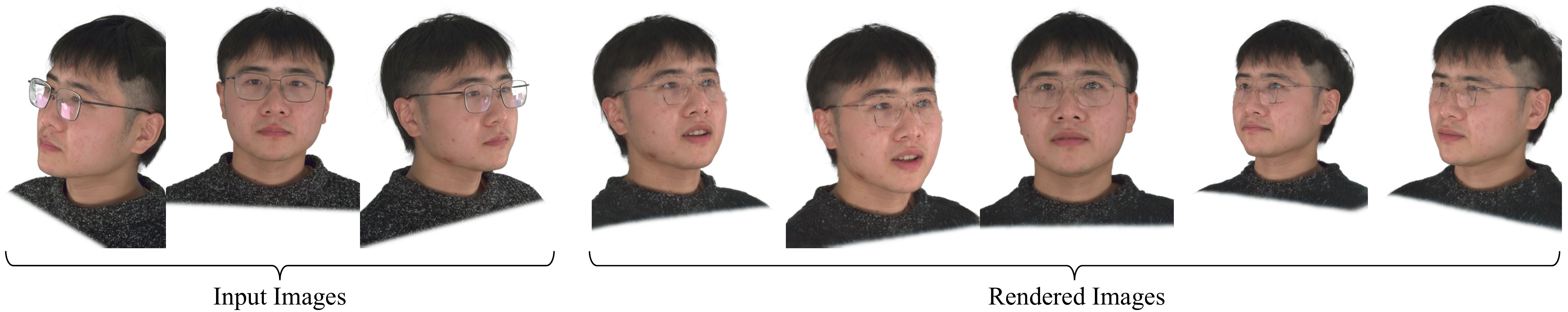}
    \caption{
    Failure cases. Our approach can not resolve subjects with noticeable facial accessories (\eg, glasses).}
    \label{fig:supp-failure}
\end{figure*}

\clearpage

\clearpage
\newpage

{
    \small
    \bibliographystyle{ieeenat_fullname}
    \bibliography{main}

\begin{thebibliography}{87}
\providecommand{\natexlab}[1]{#1}
\providecommand{\url}[1]{\texttt{#1}}
\expandafter\ifx\csname urlstyle\endcsname\relax
  \providecommand{\doi}[1]{doi: #1}\else
  \providecommand{\doi}{doi: \begingroup \urlstyle{rm}\Url}\fi

\bibitem[Athar et~al.(2022)Athar, Xu, Sunkavalli, Shechtman, and Shu]{athar2022rignerf}
ShahRukh Athar, Zexiang Xu, Kalyan Sunkavalli, Eli Shechtman, and Zhixin Shu.
\newblock Rignerf: Fully controllable neural 3d portraits.
\newblock In \emph{Proceedings of the IEEE/CVF conference on Computer Vision and Pattern Recognition}, pages 20364--20373, 2022.

\bibitem[Bai et~al.(2023)Bai, Tan, Huang, Sarkar, Tang, Qiu, Meka, Du, Dou, Orts-Escolano, et~al.]{bai2023learning}
Ziqian Bai, Feitong Tan, Zeng Huang, Kripasindhu Sarkar, Danhang Tang, Di Qiu, Abhimitra Meka, Ruofei Du, Mingsong Dou, Sergio Orts-Escolano, et~al.
\newblock Learning personalized high quality volumetric head avatars from monocular rgb videos.
\newblock In \emph{Proceedings of the IEEE/CVF Conference on Computer Vision and Pattern Recognition}, pages 16890--16900, 2023.

\bibitem[Bergman et~al.(2022)Bergman, Kellnhofer, Yifan, Chan, Lindell, and Wetzstein]{bergman2022generative}
Alexander Bergman, Petr Kellnhofer, Wang Yifan, Eric Chan, David Lindell, and Gordon Wetzstein.
\newblock Generative neural articulated radiance fields.
\newblock \emph{Advances in Neural Information Processing Systems}, 35:\penalty0 19900--19916, 2022.

\bibitem[Blanz and Vetter(2003)]{blanz2003face}
Volker Blanz and Thomas Vetter.
\newblock Face recognition based on fitting a 3d morphable model.
\newblock \emph{IEEE Transactions on pattern analysis and machine intelligence}, 25\penalty0 (9):\penalty0 1063--1074, 2003.

\bibitem[Blanz and Vetter(2023)]{blanz2023morphable}
Volker Blanz and Thomas Vetter.
\newblock A morphable model for the synthesis of 3d faces.
\newblock In \emph{Seminal Graphics Papers: Pushing the Boundaries, Volume 2}, pages 157--164. 2023.

\bibitem[Bounareli et~al.(2023)Bounareli, Tzelepis, Argyriou, Patras, and Tzimiropoulos]{bounareli2023hyperreenact}
Stella Bounareli, Christos Tzelepis, Vasileios Argyriou, Ioannis Patras, and Georgios Tzimiropoulos.
\newblock Hyperreenact: one-shot reenactment via jointly learning to refine and retarget faces.
\newblock In \emph{Proceedings of the IEEE/CVF International Conference on Computer Vision}, pages 7149--7159, 2023.

\bibitem[B{\"u}hler et~al.(2023)B{\"u}hler, Sarkar, Shah, Li, Wang, Helminger, Orts-Escolano, Lagun, Hilliges, Beeler, et~al.]{buhler2023preface}
Marcel~C B{\"u}hler, Kripasindhu Sarkar, Tanmay Shah, Gengyan Li, Daoye Wang, Leonhard Helminger, Sergio Orts-Escolano, Dmitry Lagun, Otmar Hilliges, Thabo Beeler, et~al.
\newblock Preface: A data-driven volumetric prior for few-shot ultra high-resolution face synthesis.
\newblock In \emph{Proceedings of the IEEE/CVF International Conference on Computer Vision}, pages 3402--3413, 2023.

\bibitem[Cao et~al.(2022)Cao, Simon, Kim, Schwartz, Zollhoefer, Saito, Lombardi, Wei, Belko, Yu, et~al.]{cao2022authentic}
Chen Cao, Tomas Simon, Jin~Kyu Kim, Gabriel Schwartz, Michael Zollhoefer, Shun-Suke Saito, Stephen Lombardi, Shih-En Wei, Danielle Belko, Shoou-I Yu, et~al.
\newblock Authentic volumetric avatars from a phone scan.
\newblock \emph{ACM Transactions on Graphics (TOG)}, 41\penalty0 (4):\penalty0 1--19, 2022.

\bibitem[Chan et~al.(2021)Chan, Monteiro, Kellnhofer, Wu, and Wetzstein]{chan2021pi}
Eric~R Chan, Marco Monteiro, Petr Kellnhofer, Jiajun Wu, and Gordon Wetzstein.
\newblock pi-gan: Periodic implicit generative adversarial networks for 3d-aware image synthesis.
\newblock In \emph{Proceedings of the IEEE/CVF conference on computer vision and pattern recognition}, pages 5799--5809, 2021.

\bibitem[Chan et~al.(2022)Chan, Lin, Chan, Nagano, Pan, De~Mello, Gallo, Guibas, Tremblay, Khamis, et~al.]{chan2022efficient}
Eric~R Chan, Connor~Z Lin, Matthew~A Chan, Koki Nagano, Boxiao Pan, Shalini De~Mello, Orazio Gallo, Leonidas~J Guibas, Jonathan Tremblay, Sameh Khamis, et~al.
\newblock Efficient geometry-aware 3d generative adversarial networks.
\newblock In \emph{Proceedings of the IEEE/CVF conference on computer vision and pattern recognition}, pages 16123--16133, 2022.

\bibitem[Chen et~al.(2023{\natexlab{a}})Chen, Deng, and Wang]{chen2023mimic3d}
Xingyu Chen, Yu Deng, and Baoyuan Wang.
\newblock Mimic3d: Thriving 3d-aware gans via 3d-to-2d imitation.
\newblock In \emph{2023 IEEE/CVF International Conference on Computer Vision (ICCV)}, pages 2338--2348. IEEE Computer Society, 2023{\natexlab{a}}.

\bibitem[Chen et~al.(2024)Chen, Mihajlovic, Wang, Prokudin, and Tang]{chen2024morphable}
Xiyi Chen, Marko Mihajlovic, Shaofei Wang, Sergey Prokudin, and Siyu Tang.
\newblock Morphable diffusion: 3d-consistent diffusion for single-image avatar creation.
\newblock In \emph{Proceedings of the IEEE/CVF Conference on Computer Vision and Pattern Recognition}, pages 10359--10370, 2024.

\bibitem[Chen et~al.(2023{\natexlab{b}})Chen, Wang, Li, Xiao, Zhang, Yao, and Liu]{chen2023monogaussianavatar}
Yufan Chen, Lizhen Wang, Qijing Li, Hongjiang Xiao, Shengping Zhang, Hongxun Yao, and Yebin Liu.
\newblock Monogaussianavatar: Monocular gaussian point-based head avatar.
\newblock \emph{arXiv preprint arXiv:2312.04558}, 2023{\natexlab{b}}.

\bibitem[Chu et~al.(2024)Chu, Li, Zeng, Yang, Lin, Liu, and Harada]{chu2024gpavatar}
Xuangeng Chu, Yu Li, Ailing Zeng, Tianyu Yang, Lijian Lin, Yunfei Liu, and Tatsuya Harada.
\newblock {GPA}vatar: Generalizable and precise head avatar from image(s).
\newblock In \emph{The Twelfth International Conference on Learning Representations}, 2024.

\bibitem[Deng et~al.(2019{\natexlab{a}})Deng, Guo, Xue, and Zafeiriou]{deng2019arcface}
Jiankang Deng, Jia Guo, Niannan Xue, and Stefanos Zafeiriou.
\newblock Arcface: Additive angular margin loss for deep face recognition.
\newblock In \emph{Proceedings of the IEEE/CVF conference on computer vision and pattern recognition}, pages 4690--4699, 2019{\natexlab{a}}.

\bibitem[Deng et~al.(2019{\natexlab{b}})Deng, Yang, Xu, Chen, Jia, and Tong]{deng2019accurate}
Yu Deng, Jiaolong Yang, Sicheng Xu, Dong Chen, Yunde Jia, and Xin Tong.
\newblock Accurate 3d face reconstruction with weakly-supervised learning: From single image to image set.
\newblock In \emph{Proceedings of the IEEE/CVF conference on computer vision and pattern recognition workshops}, pages 0--0, 2019{\natexlab{b}}.

\bibitem[Deng et~al.(2022)Deng, Yang, Xiang, and Tong]{deng2022gram}
Yu Deng, Jiaolong Yang, Jianfeng Xiang, and Xin Tong.
\newblock Gram: Generative radiance manifolds for 3d-aware image generation.
\newblock In \emph{Proceedings of the IEEE/CVF conference on computer vision and pattern recognition}, pages 10673--10683, 2022.

\bibitem[Deng et~al.(2023)Deng, Wang, and Shum]{deng2023learning}
Yu Deng, Baoyuan Wang, and Heung-Yeung Shum.
\newblock Learning detailed radiance manifolds for high-fidelity and 3d-consistent portrait synthesis from monocular image.
\newblock In \emph{Proceedings of the IEEE/CVF Conference on Computer Vision and Pattern Recognition}, pages 4423--4433, 2023.

\bibitem[Deng et~al.(2024{\natexlab{a}})Deng, Wang, Ren, Chen, and Wang]{deng2024portrait4d}
Yu Deng, Duomin Wang, Xiaohang Ren, Xingyu Chen, and Baoyuan Wang.
\newblock Portrait4d: Learning one-shot 4d head avatar synthesis using synthetic data.
\newblock In \emph{Proceedings of the IEEE/CVF Conference on Computer Vision and Pattern Recognition}, pages 7119--7130, 2024{\natexlab{a}}.

\bibitem[Deng et~al.(2024{\natexlab{b}})Deng, Wang, and Wang]{deng2024portrait4dv2}
Yu Deng, Duomin Wang, and Baoyuan Wang.
\newblock Portrait4d-v2: Pseudo multi-view data creates better 4d head synthesizer.
\newblock \emph{arXiv preprint arXiv:2403.13570}, 2024{\natexlab{b}}.

\bibitem[Ding et~al.(2023)Ding, Zhang, Xia, Jebe, Tu, and Zhang]{ding2023diffusionrig}
Zheng Ding, Xuaner Zhang, Zhihao Xia, Lars Jebe, Zhuowen Tu, and Xiuming Zhang.
\newblock Diffusionrig: Learning personalized priors for facial appearance editing.
\newblock In \emph{Proceedings of the IEEE/CVF Conference on Computer Vision and Pattern Recognition}, pages 12736--12746, 2023.

\bibitem[Drobyshev et~al.(2022)Drobyshev, Chelishev, Khakhulin, Ivakhnenko, Lempitsky, and Zakharov]{drobyshev2022megaportraits}
Nikita Drobyshev, Jenya Chelishev, Taras Khakhulin, Aleksei Ivakhnenko, Victor Lempitsky, and Egor Zakharov.
\newblock Megaportraits: One-shot megapixel neural head avatars.
\newblock In \emph{Proceedings of the 30th ACM International Conference on Multimedia}, pages 2663--2671, 2022.

\bibitem[Feng et~al.(2021)Feng, Feng, Black, and Bolkart]{feng2021learning}
Yao Feng, Haiwen Feng, Michael~J Black, and Timo Bolkart.
\newblock Learning an animatable detailed 3d face model from in-the-wild images.
\newblock \emph{ACM Transactions on Graphics (ToG)}, 40\penalty0 (4):\penalty0 1--13, 2021.

\bibitem[Gafni et~al.(2021)Gafni, Thies, Zollhofer, and Nie{\ss}ner]{gafni2021dynamic}
Guy Gafni, Justus Thies, Michael Zollhofer, and Matthias Nie{\ss}ner.
\newblock Dynamic neural radiance fields for monocular 4d facial avatar reconstruction.
\newblock In \emph{Proceedings of the IEEE/CVF Conference on Computer Vision and Pattern Recognition}, pages 8649--8658, 2021.

\bibitem[Gao et~al.(2022)Gao, Zhong, Xiang, Hong, Guo, and Zhang]{gao2022reconstructing}
Xuan Gao, Chenglai Zhong, Jun Xiang, Yang Hong, Yudong Guo, and Juyong Zhang.
\newblock Reconstructing personalized semantic facial nerf models from monocular video.
\newblock \emph{ACM Transactions on Graphics (TOG)}, 41\penalty0 (6):\penalty0 1--12, 2022.

\bibitem[Gao et~al.(2023)Gao, Zhou, Wang, Li, Ming, and Lu]{gao2023high}
Yue Gao, Yuan Zhou, Jinglu Wang, Xiao Li, Xiang Ming, and Yan Lu.
\newblock High-fidelity and freely controllable talking head video generation.
\newblock In \emph{Proceedings of the IEEE/CVF Conference on Computer Vision and Pattern Recognition}, pages 5609--5619, 2023.

\bibitem[Giebenhain et~al.(2023)Giebenhain, Kirschstein, Georgopoulos, R{\"u}nz, Agapito, and Nie{\ss}ner]{giebenhain2023learning}
Simon Giebenhain, Tobias Kirschstein, Markos Georgopoulos, Martin R{\"u}nz, Lourdes Agapito, and Matthias Nie{\ss}ner.
\newblock Learning neural parametric head models.
\newblock In \emph{Proceedings of the IEEE/CVF Conference on Computer Vision and Pattern Recognition}, pages 21003--21012, 2023.

\bibitem[Giebenhain et~al.(2024{\natexlab{a}})Giebenhain, Kirschstein, Georgopoulos, R{\"u}nz, Agapito, and Nie{\ss}ner]{giebenhain2024mononphm}
Simon Giebenhain, Tobias Kirschstein, Markos Georgopoulos, Martin R{\"u}nz, Lourdes Agapito, and Matthias Nie{\ss}ner.
\newblock Mononphm: Dynamic head reconstruction from monocular videos.
\newblock In \emph{Proceedings of the IEEE/CVF Conference on Computer Vision and Pattern Recognition}, pages 10747--10758, 2024{\natexlab{a}}.

\bibitem[Giebenhain et~al.(2024{\natexlab{b}})Giebenhain, Kirschstein, R{\"u}nz, Agapito, and Nie{\ss}ner]{giebenhain2024npga}
Simon Giebenhain, Tobias Kirschstein, Martin R{\"u}nz, Lourdes Agapito, and Matthias Nie{\ss}ner.
\newblock Npga: Neural parametric gaussian avatars.
\newblock \emph{arXiv preprint arXiv:2405.19331}, 2024{\natexlab{b}}.

\bibitem[Grassal et~al.(2022)Grassal, Prinzler, Leistner, Rother, Nie{\ss}ner, and Thies]{grassal2022neural}
Philip-William Grassal, Malte Prinzler, Titus Leistner, Carsten Rother, Matthias Nie{\ss}ner, and Justus Thies.
\newblock Neural head avatars from monocular rgb videos.
\newblock In \emph{Proceedings of the IEEE/CVF Conference on Computer Vision and Pattern Recognition}, pages 18653--18664, 2022.

\bibitem[Gu et~al.(2021)Gu, Liu, Wang, and Theobalt]{gu2021stylenerf}
Jiatao Gu, Lingjie Liu, Peng Wang, and Christian Theobalt.
\newblock Stylenerf: A style-based 3d-aware generator for high-resolution image synthesis.
\newblock \emph{arXiv preprint arXiv:2110.08985}, 2021.

\bibitem[Guan et~al.(2023)Guan, Zhang, Zhou, Hu, Wang, He, Feng, Liu, Ding, Liu, et~al.]{guan2023stylesync}
Jiazhi Guan, Zhanwang Zhang, Hang Zhou, Tianshu Hu, Kaisiyuan Wang, Dongliang He, Haocheng Feng, Jingtuo Liu, Errui Ding, Ziwei Liu, et~al.
\newblock Stylesync: High-fidelity generalized and personalized lip sync in style-based generator.
\newblock In \emph{Proceedings of the IEEE/CVF Conference on Computer Vision and Pattern Recognition}, pages 1505--1515, 2023.

\bibitem[Hong et~al.(2022{\natexlab{a}})Hong, Zhang, Shen, and Xu]{hong2022depth}
Fa-Ting Hong, Longhao Zhang, Li Shen, and Dan Xu.
\newblock Depth-aware generative adversarial network for talking head video generation.
\newblock In \emph{Proceedings of the IEEE/CVF conference on computer vision and pattern recognition}, pages 3397--3406, 2022{\natexlab{a}}.

\bibitem[Hong et~al.(2022{\natexlab{b}})Hong, Peng, Xiao, Liu, and Zhang]{hong2022headnerf}
Yang Hong, Bo Peng, Haiyao Xiao, Ligang Liu, and Juyong Zhang.
\newblock Headnerf: A real-time nerf-based parametric head model.
\newblock In \emph{Proceedings of the IEEE/CVF Conference on Computer Vision and Pattern Recognition}, pages 20374--20384, 2022{\natexlab{b}}.

\bibitem[Karras et~al.(2020)Karras, Laine, Aittala, Hellsten, Lehtinen, and Aila]{karras2020analyzing}
Tero Karras, Samuli Laine, Miika Aittala, Janne Hellsten, Jaakko Lehtinen, and Timo Aila.
\newblock Analyzing and improving the image quality of stylegan.
\newblock In \emph{Proceedings of the IEEE/CVF conference on computer vision and pattern recognition}, pages 8110--8119, 2020.

\bibitem[Kerbl et~al.(2023)Kerbl, Kopanas, Leimk{\"u}hler, and Drettakis]{kerbl20233d}
Bernhard Kerbl, Georgios Kopanas, Thomas Leimk{\"u}hler, and George Drettakis.
\newblock 3d gaussian splatting for real-time radiance field rendering.
\newblock \emph{ACM Transactions on Graphics}, 42\penalty0 (4):\penalty0 1--14, 2023.

\bibitem[Khakhulin et~al.(2022)Khakhulin, Sklyarova, Lempitsky, and Zakharov]{khakhulin2022realistic}
Taras Khakhulin, Vanessa Sklyarova, Victor Lempitsky, and Egor Zakharov.
\newblock Realistic one-shot mesh-based head avatars.
\newblock In \emph{European Conference on Computer Vision}, pages 345--362. Springer, 2022.

\bibitem[Kim et~al.(2018)Kim, Garrido, Tewari, Xu, Thies, Niessner, P{\'e}rez, Richardt, Zollh{\"o}fer, and Theobalt]{kim2018deep}
Hyeongwoo Kim, Pablo Garrido, Ayush Tewari, Weipeng Xu, Justus Thies, Matthias Niessner, Patrick P{\'e}rez, Christian Richardt, Michael Zollh{\"o}fer, and Christian Theobalt.
\newblock Deep video portraits.
\newblock \emph{ACM transactions on graphics (TOG)}, 37\penalty0 (4):\penalty0 1--14, 2018.

\bibitem[Kingma and Ba(2014)]{kingma2014adam}
Diederik~P Kingma and Jimmy Ba.
\newblock Adam: A method for stochastic optimization.
\newblock \emph{arXiv preprint arXiv:1412.6980}, 2014.

\bibitem[Kirschstein et~al.(2023)Kirschstein, Qian, Giebenhain, Walter, and Nie{\ss}ner]{kirschstein2023nersemble}
Tobias Kirschstein, Shenhan Qian, Simon Giebenhain, Tim Walter, and Matthias Nie{\ss}ner.
\newblock Nersemble: Multi-view radiance field reconstruction of human heads.
\newblock \emph{ACM Transactions on Graphics (TOG)}, 42\penalty0 (4):\penalty0 1--14, 2023.

\bibitem[Kirschstein et~al.(2024)Kirschstein, Giebenhain, and Nie{\ss}ner]{kirschstein2024diffusionavatars}
Tobias Kirschstein, Simon Giebenhain, and Matthias Nie{\ss}ner.
\newblock Diffusionavatars: Deferred diffusion for high-fidelity 3d head avatars.
\newblock In \emph{Proceedings of the IEEE/CVF Conference on Computer Vision and Pattern Recognition}, pages 5481--5492, 2024.

\bibitem[Ko et~al.(2023)Ko, Cho, Choi, Ryoo, and Kim]{ko20233d}
Jaehoon Ko, Kyusun Cho, Daewon Choi, Kwangrok Ryoo, and Seungryong Kim.
\newblock 3d gan inversion with pose optimization.
\newblock In \emph{Proceedings of the IEEE/CVF Winter Conference on Applications of Computer Vision}, pages 2967--2976, 2023.

\bibitem[Li et~al.(2017)Li, Bolkart, Black, Li, and Romero]{li2017learning}
Tianye Li, Timo Bolkart, Michael~J Black, Hao Li, and Javier Romero.
\newblock Learning a model of facial shape and expression from 4d scans.
\newblock \emph{ACM Trans. Graph.}, 36\penalty0 (6):\penalty0 194--1, 2017.

\bibitem[Li et~al.(2023)Li, Zhang, Wang, Zhao, Wang, Chen, Zhang, Wang, Bo, and Li]{li2023one}
Weichuang Li, Longhao Zhang, Dong Wang, Bin Zhao, Zhigang Wang, Mulin Chen, Bang Zhang, Zhongjian Wang, Liefeng Bo, and Xuelong Li.
\newblock One-shot high-fidelity talking-head synthesis with deformable neural radiance field.
\newblock In \emph{Proceedings of the IEEE/CVF Conference on Computer Vision and Pattern Recognition}, pages 17969--17978, 2023.

\bibitem[Li et~al.(2024)Li, De~Mello, Liu, Nagano, Iqbal, and Kautz]{li2024generalizable}
Xueting Li, Shalini De~Mello, Sifei Liu, Koki Nagano, Umar Iqbal, and Jan Kautz.
\newblock Generalizable one-shot 3d neural head avatar.
\newblock \emph{Advances in Neural Information Processing Systems}, 36, 2024.

\bibitem[Lombardi et~al.(2021)Lombardi, Simon, Schwartz, Zollhoefer, Sheikh, and Saragih]{lombardi2021mixture}
Stephen Lombardi, Tomas Simon, Gabriel Schwartz, Michael Zollhoefer, Yaser Sheikh, and Jason Saragih.
\newblock Mixture of volumetric primitives for efficient neural rendering.
\newblock \emph{ACM Transactions on Graphics (ToG)}, 40\penalty0 (4):\penalty0 1--13, 2021.

\bibitem[Luo et~al.(2024)Luo, Ouyang, Zhao, Jiang, Zhang, Zhang, Yang, Xu, and Yu]{luo2024gaussianhair}
Haimin Luo, Min Ouyang, Zijun Zhao, Suyi Jiang, Longwen Zhang, Qixuan Zhang, Wei Yang, Lan Xu, and Jingyi Yu.
\newblock Gaussianhair: Hair modeling and rendering with light-aware gaussians.
\newblock \emph{arXiv preprint arXiv:2402.10483}, 2024.

\bibitem[Ma et~al.(2024)Ma, Zhang, Sun, Yan, Han, and Xie]{ma2024cvthead}
Haoyu Ma, Tong Zhang, Shanlin Sun, Xiangyi Yan, Kun Han, and Xiaohui Xie.
\newblock Cvthead: One-shot controllable head avatar with vertex-feature transformer.
\newblock In \emph{Proceedings of the IEEE/CVF Winter Conference on Applications of Computer Vision}, pages 6131--6141, 2024.

\bibitem[Ma et~al.(2021)Ma, Simon, Saragih, Wang, Li, De~La~Torre, and Sheikh]{ma2021pixel}
Shugao Ma, Tomas Simon, Jason Saragih, Dawei Wang, Yuecheng Li, Fernando De~La~Torre, and Yaser Sheikh.
\newblock Pixel codec avatars.
\newblock In \emph{Proceedings of the IEEE/CVF Conference on Computer Vision and Pattern Recognition}, pages 64--73, 2021.

\bibitem[Ma et~al.(2023)Ma, Zhu, Qi, Lei, and Zhang]{ma2023otavatar}
Zhiyuan Ma, Xiangyu Zhu, Guo-Jun Qi, Zhen Lei, and Lei Zhang.
\newblock Otavatar: One-shot talking face avatar with controllable tri-plane rendering.
\newblock In \emph{Proceedings of the IEEE/CVF Conference on Computer Vision and Pattern Recognition}, pages 16901--16910, 2023.

\bibitem[Mildenhall et~al.(2021)Mildenhall, Srinivasan, Tancik, Barron, Ramamoorthi, and Ng]{mildenhall2021nerf}
Ben Mildenhall, Pratul~P Srinivasan, Matthew Tancik, Jonathan~T Barron, Ravi Ramamoorthi, and Ren Ng.
\newblock Nerf: Representing scenes as neural radiance fields for view synthesis.
\newblock \emph{Communications of the ACM}, 65\penalty0 (1):\penalty0 99--106, 2021.

\bibitem[Nguyen-Phuoc et~al.(2019)Nguyen-Phuoc, Li, Theis, Richardt, and Yang]{nguyen2019hologan}
Thu Nguyen-Phuoc, Chuan Li, Lucas Theis, Christian Richardt, and Yong-Liang Yang.
\newblock Hologan: Unsupervised learning of 3d representations from natural images.
\newblock In \emph{Proceedings of the IEEE/CVF International Conference on Computer Vision}, pages 7588--7597, 2019.

\bibitem[Paysan et~al.(2009)Paysan, Knothe, Amberg, Romdhani, and Vetter]{paysan20093d}
Pascal Paysan, Reinhard Knothe, Brian Amberg, Sami Romdhani, and Thomas Vetter.
\newblock A 3d face model for pose and illumination invariant face recognition.
\newblock In \emph{2009 sixth IEEE international conference on advanced video and signal based surveillance}, pages 296--301. Ieee, 2009.

\bibitem[Qian et~al.(2024)Qian, Kirschstein, Schoneveld, Davoli, Giebenhain, and Nie{\ss}ner]{qian2024gaussianavatars}
Shenhan Qian, Tobias Kirschstein, Liam Schoneveld, Davide Davoli, Simon Giebenhain, and Matthias Nie{\ss}ner.
\newblock Gaussianavatars: Photorealistic head avatars with rigged 3d gaussians.
\newblock In \emph{Proceedings of the IEEE/CVF Conference on Computer Vision and Pattern Recognition}, pages 20299--20309, 2024.

\bibitem[Ren et~al.(2021)Ren, Li, Chen, Li, and Liu]{ren2021pirenderer}
Yurui Ren, Ge Li, Yuanqi Chen, Thomas~H Li, and Shan Liu.
\newblock Pirenderer: Controllable portrait image generation via semantic neural rendering.
\newblock In \emph{Proceedings of the IEEE/CVF international conference on computer vision}, pages 13759--13768, 2021.

\bibitem[Roich et~al.(2022)Roich, Mokady, Bermano, and Cohen-Or]{roich2022pivotal}
Daniel Roich, Ron Mokady, Amit~H Bermano, and Daniel Cohen-Or.
\newblock Pivotal tuning for latent-based editing of real images.
\newblock \emph{ACM Transactions on graphics (TOG)}, 42\penalty0 (1):\penalty0 1--13, 2022.

\bibitem[Schwarz et~al.(2020)Schwarz, Liao, Niemeyer, and Geiger]{schwarz2020graf}
Katja Schwarz, Yiyi Liao, Michael Niemeyer, and Andreas Geiger.
\newblock Graf: Generative radiance fields for 3d-aware image synthesis.
\newblock \emph{Advances in Neural Information Processing Systems}, 33:\penalty0 20154--20166, 2020.

\bibitem[Shao et~al.(2024)Shao, Wang, Li, Wang, Lin, Zhang, Fan, and Wang]{shao2024splattingavatar}
Zhijing Shao, Zhaolong Wang, Zhuang Li, Duotun Wang, Xiangru Lin, Yu Zhang, Mingming Fan, and Zeyu Wang.
\newblock Splattingavatar: Realistic real-time human avatars with mesh-embedded gaussian splatting.
\newblock In \emph{Proceedings of the IEEE/CVF Conference on Computer Vision and Pattern Recognition}, pages 1606--1616, 2024.

\bibitem[Siarohin et~al.(2019)Siarohin, Lathuili{\`e}re, Tulyakov, Ricci, and Sebe]{siarohin2019first}
Aliaksandr Siarohin, St{\'e}phane Lathuili{\`e}re, Sergey Tulyakov, Elisa Ricci, and Nicu Sebe.
\newblock First order motion model for image animation.
\newblock \emph{Advances in neural information processing systems}, 32, 2019.

\bibitem[Sorkine and Alexa(2007)]{sorkine2007rigid}
Olga Sorkine and Marc Alexa.
\newblock As-rigid-as-possible surface modeling.
\newblock In \emph{Symposium on Geometry processing}, pages 109--116. Citeseer, 2007.

\bibitem[Sun et~al.(2022{\natexlab{a}})Sun, Wang, Shi, Wang, Wang, and Liu]{sun2022ide}
Jingxiang Sun, Xuan Wang, Yichun Shi, Lizhen Wang, Jue Wang, and Yebin Liu.
\newblock Ide-3d: Interactive disentangled editing for high-resolution 3d-aware portrait synthesis.
\newblock \emph{ACM Transactions on Graphics (ToG)}, 41\penalty0 (6):\penalty0 1--10, 2022{\natexlab{a}}.

\bibitem[Sun et~al.(2023)Sun, Wang, Wang, Li, Zhang, Zhang, and Liu]{sun2023next3d}
Jingxiang Sun, Xuan Wang, Lizhen Wang, Xiaoyu Li, Yong Zhang, Hongwen Zhang, and Yebin Liu.
\newblock Next3d: Generative neural texture rasterization for 3d-aware head avatars.
\newblock In \emph{Proceedings of the IEEE/CVF conference on computer vision and pattern recognition}, pages 20991--21002, 2023.

\bibitem[Sun et~al.(2022{\natexlab{b}})Sun, Wu, Huang, Zhang, Wang, and Li]{sun2022controllable}
Keqiang Sun, Shangzhe Wu, Zhaoyang Huang, Ning Zhang, Quan Wang, and HongSheng Li.
\newblock Controllable 3d face synthesis with conditional generative occupancy fields.
\newblock \emph{Advances in Neural Information Processing Systems}, 35:\penalty0 16331--16343, 2022{\natexlab{b}}.

\bibitem[Tang et~al.(2023)Tang, Zhang, Yang, Zhang, Chen, Ma, and Wen]{tang20233dfaceshop}
Junshu Tang, Bo Zhang, Binxin Yang, Ting Zhang, Dong Chen, Lizhuang Ma, and Fang Wen.
\newblock 3dfaceshop: Explicitly controllable 3d-aware portrait generation.
\newblock \emph{IEEE Transactions on Visualization and Computer Graphics}, 2023.

\bibitem[Tran et~al.(2024)Tran, Zakharov, Ho, Tran, Hu, and Li]{tran2024voodoo}
Phong Tran, Egor Zakharov, Long-Nhat Ho, Anh~Tuan Tran, Liwen Hu, and Hao Li.
\newblock Voodoo 3d: Volumetric portrait disentanglement for one-shot 3d head reenactment.
\newblock In \emph{Proceedings of the IEEE/CVF Conference on Computer Vision and Pattern Recognition}, pages 10336--10348, 2024.

\bibitem[Wang et~al.(2021)Wang, Mallya, and Liu]{wang2021one}
Ting-Chun Wang, Arun Mallya, and Ming-Yu Liu.
\newblock One-shot free-view neural talking-head synthesis for video conferencing.
\newblock In \emph{Proceedings of the IEEE/CVF conference on computer vision and pattern recognition}, pages 10039--10049, 2021.

\bibitem[Wiles et~al.(2018)Wiles, Koepke, and Zisserman]{wiles2018x2face}
Olivia Wiles, A Koepke, and Andrew Zisserman.
\newblock X2face: A network for controlling face generation using images, audio, and pose codes.
\newblock In \emph{Proceedings of the European conference on computer vision (ECCV)}, pages 670--686, 2018.

\bibitem[Wu et~al.(2022)Wu, Deng, Yang, Wei, Chen, and Tong]{wu2022anifacegan}
Yue Wu, Yu Deng, Jiaolong Yang, Fangyun Wei, Qifeng Chen, and Xin Tong.
\newblock Anifacegan: Animatable 3d-aware face image generation for video avatars.
\newblock \emph{Advances in Neural Information Processing Systems}, 35:\penalty0 36188--36201, 2022.

\bibitem[Wu et~al.(2023)Wu, Xu, Xiang, Wei, Chen, Yang, and Tong]{wu2023aniportraitgan}
Yue Wu, Sicheng Xu, Jianfeng Xiang, Fangyun Wei, Qifeng Chen, Jiaolong Yang, and Xin Tong.
\newblock Aniportraitgan: animatable 3d portrait generation from 2d image collections.
\newblock In \emph{SIGGRAPH Asia 2023 Conference Papers}, pages 1--9, 2023.

\bibitem[Xiang et~al.(2024)Xiang, Gao, Guo, and Zhang]{xiang2024flashavatar}
Jun Xiang, Xuan Gao, Yudong Guo, and Juyong Zhang.
\newblock Flashavatar: High-fidelity head avatar with efficient gaussian embedding.
\newblock In \emph{Proceedings of the IEEE/CVF Conference on Computer Vision and Pattern Recognition}, pages 1802--1812, 2024.

\bibitem[Xie et~al.(2023)Xie, Ouyang, Piao, Lei, and Chen]{xie2023high}
Jiaxin Xie, Hao Ouyang, Jingtan Piao, Chenyang Lei, and Qifeng Chen.
\newblock High-fidelity 3d gan inversion by pseudo-multi-view optimization.
\newblock In \emph{Proceedings of the IEEE/CVF Conference on Computer Vision and Pattern Recognition}, pages 321--331, 2023.

\bibitem[Xu et~al.(2023{\natexlab{a}})Xu, Song, Jiang, Zhang, Shi, Liu, Ma, Feng, and Luo]{xu2023omniavatar}
Hongyi Xu, Guoxian Song, Zihang Jiang, Jianfeng Zhang, Yichun Shi, Jing Liu, Wanchun Ma, Jiashi Feng, and Linjie Luo.
\newblock Omniavatar: Geometry-guided controllable 3d head synthesis.
\newblock In \emph{Proceedings of the IEEE/CVF Conference on Computer Vision and Pattern Recognition}, pages 12814--12824, 2023{\natexlab{a}}.

\bibitem[Xu et~al.(2023{\natexlab{b}})Xu, Wang, Zhao, Zhang, and Liu]{xu2023avatarmav}
Yuelang Xu, Lizhen Wang, Xiaochen Zhao, Hongwen Zhang, and Yebin Liu.
\newblock Avatarmav: Fast 3d head avatar reconstruction using motion-aware neural voxels.
\newblock In \emph{ACM SIGGRAPH 2023 Conference Proceedings}, pages 1--10, 2023{\natexlab{b}}.

\bibitem[Xu et~al.(2023{\natexlab{c}})Xu, Zhang, Wang, Zhao, Huang, Qi, and Liu]{xu2023latentavatar}
Yuelang Xu, Hongwen Zhang, Lizhen Wang, Xiaochen Zhao, Han Huang, Guojun Qi, and Yebin Liu.
\newblock Latentavatar: Learning latent expression code for expressive neural head avatar.
\newblock In \emph{ACM SIGGRAPH 2023 Conference Proceedings}, pages 1--10, 2023{\natexlab{c}}.

\bibitem[Xu et~al.(2024)Xu, Chen, Li, Zhang, Wang, Zheng, and Liu]{xu2024gaussian}
Yuelang Xu, Benwang Chen, Zhe Li, Hongwen Zhang, Lizhen Wang, Zerong Zheng, and Yebin Liu.
\newblock Gaussian head avatar: Ultra high-fidelity head avatar via dynamic gaussians.
\newblock In \emph{Proceedings of the IEEE/CVF Conference on Computer Vision and Pattern Recognition}, pages 1931--1941, 2024.

\bibitem[Yang et~al.(2024)Yang, Zheng, Ma, Lai, Wan, and Huang]{yang2024vrmm}
Haotian Yang, Mingwu Zheng, Chongyang Ma, Yu-Kun Lai, Pengfei Wan, and Haibin Huang.
\newblock Vrmm: A volumetric relightable morphable head model.
\newblock \emph{arXiv preprint arXiv:2402.04101}, 2024.

\bibitem[Ye et~al.(2024)Ye, Zhong, Ren, Yang, Li, Huang, Jiang, He, Huang, Liu, et~al.]{ye2024real3d}
Zhenhui Ye, Tianyun Zhong, Yi Ren, Jiaqi Yang, Weichuang Li, Jiawei Huang, Ziyue Jiang, Jinzheng He, Rongjie Huang, Jinglin Liu, et~al.
\newblock Real3d-portrait: One-shot realistic 3d talking portrait synthesis.
\newblock \emph{arXiv preprint arXiv:2401.08503}, 2024.

\bibitem[Yin et~al.(2022)Yin, Zhang, Cun, Cao, Fan, Wang, Bai, Wu, Wang, and Yang]{yin2022styleheat}
Fei Yin, Yong Zhang, Xiaodong Cun, Mingdeng Cao, Yanbo Fan, Xuan Wang, Qingyan Bai, Baoyuan Wu, Jue Wang, and Yujiu Yang.
\newblock Styleheat: One-shot high-resolution editable talking face generation via pre-trained stylegan.
\newblock In \emph{European conference on computer vision}, pages 85--101. Springer, 2022.

\bibitem[Yin et~al.(2023)Yin, Ghasedi, Wu, Yang, Tong, and Fu]{yin2023nerfinvertor}
Yu Yin, Kamran Ghasedi, HsiangTao Wu, Jiaolong Yang, Xin Tong, and Yun Fu.
\newblock Nerfinvertor: High fidelity nerf-gan inversion for single-shot real image animation.
\newblock In \emph{Proceedings of the IEEE/CVF Conference on Computer Vision and Pattern Recognition}, pages 8539--8548, 2023.

\bibitem[Yu et~al.(2023)Yu, Fan, Zhang, Wang, Yin, Bai, Cao, Shan, Wu, Sun, et~al.]{yu2023nofa}
Wangbo Yu, Yanbo Fan, Yong Zhang, Xuan Wang, Fei Yin, Yunpeng Bai, Yan-Pei Cao, Ying Shan, Yang Wu, Zhongqian Sun, et~al.
\newblock Nofa: Nerf-based one-shot facial avatar reconstruction.
\newblock In \emph{ACM SIGGRAPH 2023 Conference Proceedings}, pages 1--12, 2023.

\bibitem[Yu et~al.(2024)Yu, Bai, Meka, Tan, Xu, Pandey, Fanello, Park, and Zhang]{yu2024one2avatar}
Zhixuan Yu, Ziqian Bai, Abhimitra Meka, Feitong Tan, Qiangeng Xu, Rohit Pandey, Sean Fanello, Hyun~Soo Park, and Yinda Zhang.
\newblock One2avatar: Generative implicit head avatar for few-shot user adaptation.
\newblock \emph{arXiv preprint arXiv:2402.11909}, 2024.

\bibitem[Zhang et~al.(2023)Zhang, Qi, Zhang, Zhang, Wu, Chen, Chen, Wang, and Wen]{zhang2023metaportrait}
Bowen Zhang, Chenyang Qi, Pan Zhang, Bo Zhang, HsiangTao Wu, Dong Chen, Qifeng Chen, Yong Wang, and Fang Wen.
\newblock Metaportrait: Identity-preserving talking head generation with fast personalized adaptation.
\newblock In \emph{Proceedings of the IEEE/CVF Conference on Computer Vision and Pattern Recognition}, pages 22096--22105, 2023.

\bibitem[Zheng et~al.(2024)Zheng, Wen, Su, Xu, Li, Zhao, and Xue]{zheng2024ohta}
Xiaozheng Zheng, Chao Wen, Zhuo Su, Zeran Xu, Zhaohu Li, Yang Zhao, and Zhou Xue.
\newblock Ohta: One-shot hand avatar via data-driven implicit priors.
\newblock In \emph{Proceedings of the IEEE/CVF International Conference on Computer Vision}, 2024.

\bibitem[Zheng et~al.(2022)Zheng, Abrevaya, B{\"u}hler, Chen, Black, and Hilliges]{zheng2022avatar}
Yufeng Zheng, Victoria~Fern{\'a}ndez Abrevaya, Marcel~C B{\"u}hler, Xu Chen, Michael~J Black, and Otmar Hilliges.
\newblock Im avatar: Implicit morphable head avatars from videos.
\newblock In \emph{Proceedings of the IEEE/CVF Conference on Computer Vision and Pattern Recognition}, pages 13545--13555, 2022.

\bibitem[Zheng et~al.(2023)Zheng, Yifan, Wetzstein, Black, and Hilliges]{zheng2023pointavatar}
Yufeng Zheng, Wang Yifan, Gordon Wetzstein, Michael~J Black, and Otmar Hilliges.
\newblock Pointavatar: Deformable point-based head avatars from videos.
\newblock In \emph{Proceedings of the IEEE/CVF conference on computer vision and pattern recognition}, pages 21057--21067, 2023.

\bibitem[Zielonka et~al.(2022)Zielonka, Bolkart, and Thies]{zielonka2022towards}
Wojciech Zielonka, Timo Bolkart, and Justus Thies.
\newblock Towards metrical reconstruction of human faces.
\newblock In \emph{European conference on computer vision}, pages 250--269. Springer, 2022.

\bibitem[Zielonka et~al.(2023)Zielonka, Bolkart, and Thies]{zielonka2023instant}
Wojciech Zielonka, Timo Bolkart, and Justus Thies.
\newblock Instant volumetric head avatars.
\newblock In \emph{Proceedings of the IEEE/CVF Conference on Computer Vision and Pattern Recognition}, pages 4574--4584, 2023.

\end{thebibliography}
}

\end{document}